\documentclass[letterpaper]{article}
\def\SSTGCameraReady{}
\def\SSTGDocumentClassLoaded{}
\ifdefined\SSTGDocumentClassLoaded
\else
\documentclass[letterpaper]{article}
\fi
\ifdefined\SSTGCameraReady
\usepackage[preprint]{aaai2027}
\else
\usepackage[submission]{aaai2027}
\fi
\usepackage[hyphens]{url}
\usepackage{graphicx}
\def\UrlFont{\rm}
\usepackage{natbib}
\usepackage{caption}
\usepackage{booktabs}
\usepackage{multirow}
\usepackage{amsmath}
\usepackage{amssymb}
\usepackage{pifont}
\usepackage{algorithm}
\usepackage{algpseudocode}
\ifdefined\SSTGCameraReady
\else
\fi

\newcommand{\cmark}{\ding{51}}
\newcommand{\xmark}{\ding{55}}
\newcommand{\sys}{SSTG-Nav}

\title{SSTG-Nav: Metric-Grounded Spatial-Semantic Topological Graphs for Reusable Object Navigation}
\ifdefined\SSTGCameraReady
\author{
Daojie Peng\textsuperscript{\rm 1},
Bingtao Wang\textsuperscript{\rm 2},
Jun Ma\textsuperscript{\rm 1}\corresponding
}
\affiliations{
\textsuperscript{\rm 1}The Hong Kong University of Science and Technology (Guangzhou)\\
\textsuperscript{\rm 2}Shandong University\\
dpeng108@connect.hkust-gz.edu.cn, wangbt@mail.sdu.edu.cn, jun.ma@ust.hk
}
\else
\author{Anonymous AAAI Submission}
\affiliations{Anonymous Institution}
\fi

\begin{document}
\maketitle

\begin{abstract}
Service robots operating for months in the same homes, offices, and facilities should become more reliable with experience instead of searching familiar space from scratch for every request. Yet ObjectNav is predominantly formulated as one-shot exploration, leaving a central deployment challenge unresolved: recognizing an object does not identify a reachable place to stop, and one confident map error can terminate the task. We introduce \sys, a reusable metric-semantic memory that turns a one-time survey into actionable object goals, consolidates evidence across viewpoints, and retains spatially distinct recovery standoffs. On 1,000 HM3D-v2 episodes across 36 scenes, our goal-independent topology achieves a 99.4\% geometric success ceiling. Holding semantic responses fixed, metric grounding raises SR/SPL from 0.835/0.560 to 0.920/0.603, and source-aware fusion reaches 0.926/0.586. Fusion-aware Top-3 recovery raises Success@1/2/3 to 0.928/0.965/0.975 and reaches 0.601 SPL@3. Model, field-of-view, density, and corruption controls identify where these gains originate, and a ROS~2/Nav2 realization demonstrates the complete reusable query-to-execution pipeline. Together, the results establish pre-exploration as a powerful practical regime for dependable, repeated semantic navigation.
\end{abstract}

\ifdefined\SSTGCameraReady
\begin{links}
    \link{Code}{https://github.com/DaojiePENG/sstg-nav-bench}
    \link{Project}{https://daojiepeng.github.io/SSTG-Nav}
\end{links}
\fi

\section{Introduction}

ObjectNav requires an embodied agent to reach an instance of a requested category and declare STOP at a valid location \cite{long2024instructnav,peng2025lovon}. It is a foundational capability for service robots \cite{batra2020objectnav}. Most benchmarks deliberately begin in an unexplored environment, and semantic exploration \cite{chaplot2020semexp}, vision-language frontier maps \cite{yokoyama2024vlfm}, commonsense constraints \cite{zhou2023esc}, online scene graphs \cite{yin2024sgnav}, and topological memories \cite{liu2025toponav} have substantially advanced that setting. Long-lived robots, however, revisit the same homes, offices, hospitals, and facilities across hundreds of requests. Repeating full discovery wastes prior experience and makes task completion less predictable. We study the complementary, practically central question of how a robot can convert one survey into reliable navigation infrastructure for all subsequent requests.

This reusable regime changes the objective from one-episode exploration efficiency to dependable completion over a robot's lifetime. A persistent map can amortize expensive perception, answer future language queries immediately, retain several hypotheses for ambiguous objects, and recover from a rejected candidate without restarting search. The central scientific challenge is making that memory \emph{actionable}: it must encode where the robot can successfully stop, not merely what its cameras once recognized.

This distinction exposes a fundamental observation-to-action gap. A camera may recognize an object through a doorway, across a railing, or from the wrong side of furniture, while the camera pose itself is a poor navigation goal. Moreover, maps sampled from official target viewpoints already contain successful STOP configurations; we call this \emph{target-view coverage} and use it as a controlled ceiling. Our goal-independent setting addresses the harder and more useful problem: discovering navigable space without task goals, then converting visual evidence into object-centric destinations that are reachable, corroborated, and verifiable at arrival.

\begin{figure*}[t]
  \centering
  \includegraphics[width=0.98\textwidth]{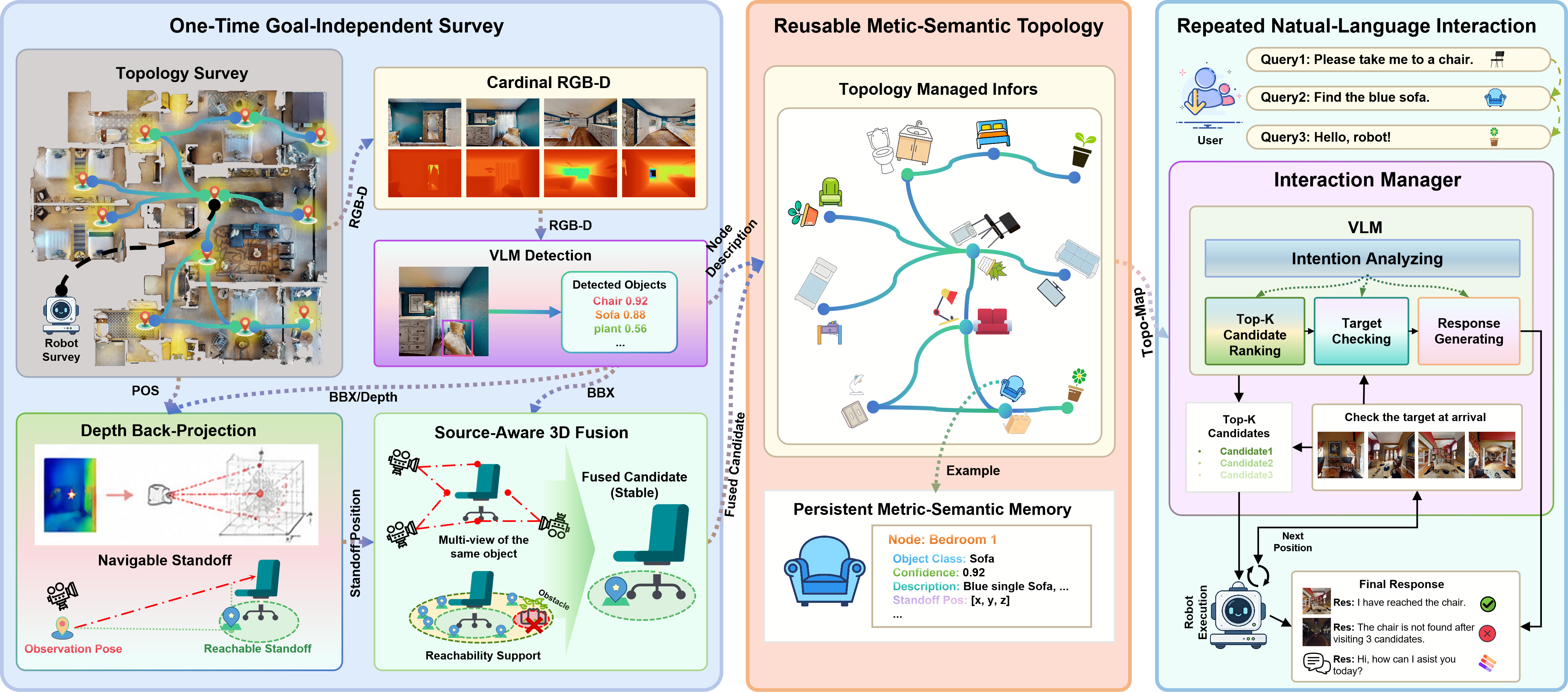}
  \caption{\sys{} converts a one-time, goal-independent survey into a
persistent metric-semantic topology that supports repeated natural-language
navigation. View-local RGB-D detections are projected into reachable
object-centric standoffs and consolidated across viewpoints. For each user
request, the interaction manager constructs a structured semantic query,
retrieves and ranks multiple candidates, coordinates graph planning and
Nav2 execution, and verifies the target from fresh arrival RGB-D. A rejected
hypothesis triggers navigation to the next candidate, while clarification,
progress, recovery, and verified completion are reported through the
bidirectional natural-language interface. 
}
  \label{fig:overview}
\end{figure*}

Figure~\ref{fig:overview} presents our solution. \sys{} separates a one-time, goal-independent survey from lightweight repeated queries. It converts view-local VLM evidence into reachable object-centric destinations, strengthens them through independent multi-view support, ranks alternatives on a sparse navigation graph, and uses fresh RGB-D evidence to decide whether to STOP or continue. Goal annotations terminate in a separate post-hoc evaluator lane. This design preserves the efficiency of topological memory while supplying the metric grounding and closed-loop recovery needed for dependable execution.

Our contributions are:
\begin{itemize}
  \item \textbf{Scientific formulation:} we isolate reusable ObjectNav as an operating regime and identify the observation-pose/STOP-pose mismatch, together with protocols that separate target-derived coverage from goal-independent mapping.
  \item \textbf{Technical framework:} \sys{} turns view-local VLM detections into reachable object-centric standoffs, consolidates them through source-aware 3D fusion, and verifies candidate arrival from fresh RGB-D before STOP.
  \item \textbf{Empirical evidence:} full-validation coverage controls and same-response independent-map ablations distinguish geometry, semantic backend, FoV, metric grounding, fusion, and sequential recovery.
  \item \textbf{System realization:} a modular natural-language-to-map-to-navigation stack instantiates the interfaces in ROS~2, with closed-loop Nav2 target dispatch and physical motion verified on a mobile robot.
\end{itemize}

Together, the contributions turn pre-exploration from a favorable assumption into a rigorously evaluated navigation paradigm. Tables~\ref{tab:coverage}, \ref{tab:main}, and \ref{tab:topk} show that a task-independent map nearly saturates geometric coverage and that \sys{} converts this potential into 0.926 one-shot SR and 0.975 fusion-aware Top-3 SR. This positions reusable metric-semantic topology as a strong foundation for long-lived robots that must answer many navigation requests in stable environments.

\section{Related Work}

\textbf{Online ObjectNav.}
Early modular agents learn where to explore through semantic policies or potential maps \cite{chaplot2020semexp,ramakrishnan2022poni}, while ZSON and CoW broaden recognition toward open-world or language-driven goals \cite{majumdar2022zson,gadre2023cow}. Foundation-model systems subsequently connect language and pixel evidence to online exploration through LLM reasoning, pixel goals, value maps, commonsense constraints, or sparse Voronoi structure \cite{yu2023l3mvn,cai2024pixnav,zhou2023esc,yokoyama2024vlfm,wu2024voronav,peng2026structured}. Recent methods add generative priors, geometric affordances, scene graphs, Bayesian structure, or predictive world models \cite{zhang2024imagine,yuan2024gamap,yin2024sgnav,zhang2025fbn,yin2025unigoal,nie2025wmnav}, and increasingly organize reasoning, experience retrieval, instruction interpretation, target fusion, frontier planning, topology, learned behavior, and execution consistency \cite{cao2025cognav,wang2026trajrag,long2024instructnav,zhang2025apexnav,chabal2025fomnav,liu2025toponav,cai2026intentnav,wang2026consistnav}. These systems primarily search an initially unknown scene. \sys{} instead assumes a goal-independent prior survey and isolates the metricization, fusion, and recovery errors that remain after exploration has been amortized.

\textbf{Reusable semantic maps.}
Reusable abstract models first demonstrated that scene knowledge accumulated online can improve later ObjectNav requests \cite{campari2022reusable}. VLMaps fuses language features with 3D reconstruction for natural-language map queries \cite{huang2023vlmaps}, while ConceptFusion, OpenMask3D, and CLIP-Fields provide complementary open-set 3D or continuous semantic memories \cite{jatavallabhula2023conceptfusion,takmaz2023openmask3d,shafiullah2023clipfields}. CARe explicitly studies pre-explored maps and uses confidence and multi-view consistency to revise decisions \cite{ko2024care}. Topological graph memories compress experience into landmarks and connectivity \cite{kim2022tsgm}, and TopoNav further establishes topology as an effective substrate for advanced ObjectNav reasoning \cite{liu2025toponav}. Our representation is deliberately lighter than a dense feature field: it stores navigation nodes plus metric object candidates, then evaluates the precise conversion from a recognized observation to a valid ObjectNav STOP pose.

\section{Problem and Protocols}

Let a pre-explored scene contain navigable space $\mathcal{X}$ and a graph $\mathcal{G}=(\mathcal{V},\mathcal{E})$. A navigation node $v_i=(\mathbf{p}_i,\mathbf{q}_i,\mathcal{I}_i)$ stores its 3D position, orientation, and RGB-D views. An edge connects locally reachable nodes and is weighted by navmesh geodesic distance. A query supplies category $c$ and start $\mathbf{x}_0$. The system returns a ranked metric candidate $\hat{\mathbf{s}}$ and the shortest feasible path from $\mathbf{x}_0$ to that point.

For an official episode with valid goal-viewpoint set $\mathcal{Y}_c$, success follows the ObjectNav task and challenge protocol \cite{batra2020objectnav,habitatchallenge2023}:
\begin{equation}
S=\mathbf{1}\!\left[\min_{\mathbf{y}\in\mathcal{Y}_c}d_{\mathrm{geo}}(\hat{\mathbf{s}},\mathbf{y})\leq 1\,\mathrm{m}\right].
\end{equation}
SPL is $S\ell/\max(\ell,p)$, where $\ell$ is the official shortest-path distance and $p$ is the planned route length \cite{anderson2018evaluation}. DTG is final geodesic distance to the nearest valid viewpoint.

We use three protocols. \emph{Target-view coverage} selects one high-IoU navigable viewpoint per annotated object instance and measures a coverage-controlled upper bound. \emph{Independent topology oracle} samples the map without goals, then assigns oracle semantic labels only for evaluation. It isolates geometric coverage. \emph{Independent topology with real semantics} keeps goals hidden throughout capture, VLM inference, projection, fusion, and retrieval. This is our main protocol. Reporting the regimes separately makes the benefit of reusable mapping directly interpretable under each information budget.

\begin{table}[H]
\centering
\caption{Information boundary of the three evaluation protocols. ``Eval.'' means annotations are exposed only after mapping and candidate selection.}
\label{tab:protocols}
\resizebox{\columnwidth}{!}{\begin{tabular}{lccc}
\toprule
Protocol & Goals in map & Oracle semantics & Real VLM \\
\midrule
Target-view coverage & \cmark & optional & optional \\
Independent oracle & \xmark & Eval. & \xmark \\
Independent real (ours) & \xmark & \xmark & \cmark \\
\bottomrule
\end{tabular}}
\end{table}

Table~\ref{tab:protocols} makes the decisive distinction explicit: only the coverage control uses official goal viewpoints during map construction. In the independent real protocol, episode starts, categories, object annotations, and valid goal viewpoints remain hidden throughout topology construction, RGB-D capture, VLM inference, projection, fusion, and retrieval; they enter only after candidate selection for success and DTG evaluation. The supplement provides the complete construction audit.

\section{Method}

\subsection{Goal-Independent Spatial Topology}

For each scene, we draw a fixed-seed pool of navigable points and apply greedy farthest-point sampling until the empirical pool-cover radius is at most $r$. Edges connect up to eight nearby nodes when a navmesh path exists and its geodesic length passes a local threshold. This sampling is independent of episode starts, categories, object annotations, and goal viewpoints. The graph may be built once and reused.

At every node, the robot captures four cardinal RGB-D views. We process views separately rather than asking a VLM to localize objects in a stitched panorama, which preserves each camera model and avoids ambiguous panorama coordinates. Each returned detection is $d=(c,b,q,k)$: category, normalized box, confidence, and view index.

\subsection{From Detection to Navigable Candidate}

Let $\tilde{\mathbf{u}}$ be the box center and $z$ the median valid depth in a $7\!\times\!7$ patch. With intrinsics $K$ and world-from-camera transform $T_{wk}$, the estimated surface point is
\begin{equation}
\hat{\mathbf{o}}_d=T_{wk}\left(zK^{-1}[\tilde{\mathbf{u}}^\top,1]^\top\right).
\end{equation}
The observation pose itself can be a poor destination. We therefore move $\rho$ meters from the surface toward the source camera in the ground plane,
\begin{equation}
\mathbf{s}^{*}_d=\hat{\mathbf{o}}_d+\rho\frac{\Pi(\mathbf{p}_i-\hat{\mathbf{o}}_d)}{\|\Pi(\mathbf{p}_i-\hat{\mathbf{o}}_d)\|_2},
\end{equation}
where $\Pi$ drops the vertical coordinate. We snap $\mathbf{s}^{*}_d$ to navigable space and discard it if no navmesh path exists from the source node. This operation uses depth only for geometry. Semantic confidence remains the VLM output.

\subsection{Reachability-Aware Soft Fusion}

Detections of category $c$ form a cluster $C$ only if their object estimates are within horizontal radius $r_o$ and vertical tolerance $r_h$, and their STOP candidates are mutually reachable within geodesic threshold $r_g$. The last condition prevents Euclidean fusion through walls or between disconnected floor regions. Multiple boxes from one source image must not inflate confidence. We retain the maximum confidence per source topology node and compute a noisy-OR score
\begin{equation}
Q(C)=1-\prod_{i\in U(C)}\left(1-\max_{d\in C_i}q_d\right),
\end{equation}
where $U(C)$ is the set of independent observing nodes. A cluster is retained with two-node support or a high-confidence singleton; its highest-confidence reachable standoff becomes the representative. We store two linked layers: $\mathcal H^F$ contains one $Q$-scored representative per retained cluster, while $\mathcal H^R$ keeps all reachable pre-fusion standoffs with view-local confidence. Fused-only Top-$K$ draws every visit from $\mathcal H^F$; our policy draws visit one from $\mathcal H^F$ and spatially separated recovery visits from $\mathcal H^R$. Thus stable primary ranking does not erase alternative stopping geometry or filtered detections.

\begin{algorithm}[t]
\caption{Goal-independent metric semantic mapping}
\label{alg:mapping}
\begin{algorithmic}[1]
\Require Graph $\mathcal{G}$, RGB-D views, VLM $f$, standoff $\rho$
\State $\mathcal{H}\gets\emptyset$
\For{$v_i\in\mathcal{V}$ and view $k$}
  \For{$d=(c,b,q,k)\in f(I_{ik})$}
    \State $z\gets\Call{PatchMedian}{D_{ik},b}$
    \State $\hat{\mathbf{o}}_d\gets\Call{BackProject}{b,z,K,T_{ik}}$
    \State $\mathbf{s}^{*}_d\gets\Call{Standoff}{\hat{\mathbf{o}}_d,\mathbf{p}_i,\rho}$
    \State $\mathbf{s}_d\gets\Call{NavSnap}{\mathbf{s}^{*}_d}$
    \If{$\mathbf{s}_d$ is reachable from $v_i$}
      \State append $(c,q,\hat{\mathbf{o}}_d,\mathbf{s}_d,i)$ to $\mathcal{H}$
    \EndIf
  \EndFor
\EndFor
\For{category $c$}
  \State cluster $\mathcal{H}_c$ by 3D proximity and STOP reachability
  \State score clusters with independent-source noisy-OR
  \State retain multi-source clusters or confident singletons
  \State retain reachable residual standoffs for recovery
\EndFor
\State \Return $\mathcal{G}$ with fused representatives and residual standoffs
\end{algorithmic}
\end{algorithm}

Algorithm~\ref{alg:mapping} summarizes the complete offline transformation from calibrated observations to reusable metric candidates; no query or episode field appears in its inputs.

Sequential candidates improve fault tolerance without consulting the evaluator. At candidate $k$, the robot captures four fresh $120^\circ$ RGB-D views: view zero faces the stored object estimate and the others rotate by $90^\circ$. GPT-5.4 receives these views plus the boxed mapping-time reference and returns target visibility, a stopping-side judgment $g_k$, an arrival-view box, and confidence $q_k$. The box indexes aligned depth with central-patch median $z_k$. We accept STOP exactly when
\begin{equation}
\begin{aligned}
A_k=\mathbf{1}[&\mathrm{visible}_k\wedge g_k=\mathrm{valid}\wedge q_k\geq 0.75\\
               &\wedge 0.25\,\mathrm{m}\leq z_k\leq2.5\,\mathrm{m}].
\end{aligned}
\label{eq:arrival}
\end{equation}
On rejection, planning continues from the current pose to the next spatially diverse candidate. Official goal viewpoints enter only after this decision to score success, SPL, and verifier confusion.

\subsection{Reusable Query and Execution Layer}

Let the language normalizer map query $q$ to category $c(q)$, and let
\begin{equation}
\mathcal{C}(q)=\{i:c_i=c(q),\;\mathbf{s}_i\text{ is reachable from }\mathbf{x}_0\}
\end{equation}
be its fused primary candidates from $\mathcal H^F$. Candidate order is the lexicographic permutation
\begin{equation}
\boldsymbol{\pi}_q=\operatorname{LexSort}_{i\in\mathcal{C}(q)}
\left(-Q_i,\ d_{\mathcal G}(\mathbf{x}_0,\mathbf{s}_i)\right),
\label{eq:query-rank}
\end{equation}
which makes semantic support primary and path length a deterministic tie-breaker. This choice prevents a nearby unsupported observation from outranking independently corroborated evidence while avoiding a learned query-time policy.

Visit one uses $\pi_q(1)$. Up to two later visits come from the category-matched $\mathcal R(q)\subset\mathcal H^R$, ordered by confidence and path distance while enforcing $\|\mathbf{s}_i-\mathbf{s}_{\pi_q(j)}\|_2\geq2$\,m for every prior visit. If $\mathcal C(q)$ is empty, the best reachable residual becomes visit one. Hence fused-only tries three cluster representatives, whereas fusion-aware tries one supported representative and two view-specific standoffs.

Execution is a receding candidate sequence rather than independent start-to-goal trials. With $\mathbf{x}_0$ the request start, leg $k$ is
\begin{equation}
P_k=\operatorname{ShortestPath}(\mathcal G,\mathbf{x}_{k-1},\mathbf{s}_{\pi_q(k)}),
\qquad \mathbf{x}_k=\mathbf{s}_{\pi_q(k)},
\end{equation}
and the executed path is $P(q)=P_1\oplus\cdots\oplus P_{\tau}$ for $\tau=\min\{k:A_k=1\}$. Hence a rejected hypothesis changes the next planning state instead of restarting the episode. For $m_c=|\mathcal C(q)|$, ordering costs $O(m_c\log m_c)$ and each graph search costs $O(|\mathcal E|+|\mathcal V|\log|\mathcal V|)$. If one survey serves $N$ requests, its amortized non-motion cost is
\begin{equation}
\bar C_N=C_{\mathrm{map}}/N+C_{\mathrm{retrieve}}+C_{\mathrm{plan}},
\end{equation}
formalizing the intended advantage: VLM inference and depth grounding are paid once, while repeated requests perform retrieval, graph search, and fresh arrival verification only.

\section{Experiments}

\subsection{Setup}

We evaluate the HM3D-Semantics v0.2 ObjectNav episodes from the 2023 Habitat Navigation Challenge \cite{habitatchallenge2023,yadav2023hm3dsem}. The episodes use semantic annotations over HM3D scenes \cite{ramakrishnan2021hm3d} and are rendered in Habitat-Sim 0.3.3 \cite{savva2019habitat}. Full validation has 1,000 episodes across 36 scenes and six categories: chair, bed, plant, toilet, TV/monitor, and sofa. Goal-independent farthest-point sampling draws 12,000 navigable proposals per scene and stops at $r=0.8$\,m coverage, producing 6,642 nodes and 21,845 edges. Each node stores four $640\!\times\!640$, $120^\circ$ RGB-D views at 1.25\,m camera height. Controlled representation and model ablations use the official 30-episode, two-scene minival subset on a 339-node graph; these scales are identified explicitly in Tables~\ref{tab:coverage}--\ref{tab:topk}.

The primary semantic backend is GPT-5.4 \cite{openai2026gpt54}; MiMo-v2.5 \cite{mimov25} and local Qwen2.5-VL-3B-Instruct \cite{bai2025qwen25vl} are auxiliary model controls. We use $\rho=0.8$\,m, $r_o=1.2$\,m, $r_h=1.0$\,m, $r_g=3.0$\,m, and residual separation $\delta=2.0$\,m. Full fusion retains two-source clusters or singletons above 0.92; the small-split representation controls retain every valid cluster so category scarcity is not confounded with filtering. Arrival verification uses four fresh $640\!\times\!640$, $120^\circ$ views, confidence 0.75, and measured depth range $[0.25,2.5]$\,m. We report Wilson 95\% intervals for SR and 10,000-sample non-parametric bootstrap intervals for SPL. Supplementary Sections~1--2 audit the construction, prompts, and parameters used by all result tables.

\subsection{Comparison Across Operating Regimes}

\begin{table*}[t]
\centering
\caption{ObjectNav context on HM3D-v2. Unknown-scene methods explore per episode, whereas reusable-map methods amortize a survey across repeated queries. \sys{} achieves the highest SR and SPL among the listed results. TF denotes no task-specific policy training; CARe reports a custom MP3D success metric ($\dagger$).}
\label{tab:context}
\scriptsize
\setlength{\tabcolsep}{4.0pt}
\begin{tabular*}{\textwidth}{@{\extracolsep{\fill}}lllcrr}
\toprule
Method & Scene at start & Evaluation & Policy & SR & SPL \\
\midrule
\multicolumn{6}{l}{\emph{Unknown-scene HM3D-v2}}\\
SG-Nav \cite{yin2024sgnav,wang2026trajrag} & unknown & HM3D-v2 & TF & 0.496 & 0.255 \\
InstructNav \cite{long2024instructnav} & unknown & HM3D-v2 & TF & 0.580 & 0.209 \\
VLFM \cite{yokoyama2024vlfm,wang2026trajrag} & unknown & HM3D-v2 & TF & 0.636 & 0.325 \\
ApexNav \cite{zhang2025apexnav} & unknown & HM3D-v2 & TF & 0.762 & 0.380 \\
FOM-Nav \cite{chabal2025fomnav} & unknown & HM3D-v2 & learned & 0.758 & 0.479 \\
TrajRAG \cite{wang2026trajrag} & unknown & HM3D-v2 & TF & 0.781 & 0.402 \\
IntentNav \cite{cai2026intentnav} & unknown & HM3D-v2 & learned & 0.822 & 0.385 \\
ConsistNav \cite{wang2026consistnav} & unknown & HM3D-v2 & TF & 0.842 & 0.412 \\
\midrule
\multicolumn{6}{l}{\emph{Pre-explored reusable maps}}\\
CARe+VLMaps \cite{ko2024care} & pre-explored & custom MP3D & TF & 0.827$^\dagger$ & -- \\
\sys{} fused, single & pre-explored & HM3D-v2 & TF & 0.926 & 0.586 \\
\sys{} fusion-aware Top-3 & pre-explored & HM3D-v2 & TF & \textbf{0.975} & \textbf{0.601} \\
\bottomrule
\end{tabular*}
\end{table*}

The central result is that a reusable survey moves ObjectNav to a substantially stronger operating point. Among the unknown-scene HM3D-v2 systems listed in Table~\ref{tab:context}, the best SR and SPL are 0.842 and 0.479. A single fused \sys{} destination already reaches 0.926/0.586, and fusion-aware recovery reaches 0.975/0.601, margins of 0.133 SR and 0.122 SPL over those strongest listed values. More importantly, this reliability does not require a task-specific navigation policy: one goal-independent survey builds metric-semantic infrastructure that can serve many later requests. In recurring environments such as homes, offices, and care facilities, perception and mapping are paid once; subsequent natural-language queries reduce to retrieval, graph planning, and targeted recovery. The analyses below isolate how coverage, grounding, fusion, and residual candidates produce this advantage. Complete HM3D-v1/v2 context appears in the supplement.

\subsection{Coverage Is Necessary but Not Sufficient}

\begin{table}[t]
\centering
\caption{Full-validation coverage--semantics decomposition over 1,000 episodes. Target-view maps use goal-derived capture poses; independent maps do not.}
\label{tab:coverage}
\resizebox{\columnwidth}{!}{\begin{tabular}{llrrrr}
\toprule
Map protocol & Semantics & Nodes & SR & SPL & DTG \\
\midrule
Target-view & Oracle & 1,168 & 0.990 & 0.802 & \textbf{0.073} \\
Target-view & Qwen & 1,168 & 0.644 & 0.438 & 2.016 \\
Independent & Oracle & 6,642 & \textbf{0.994} & \textbf{0.992} & 0.622 \\
Independent & GPT-5.4 fused & 6,642 & 0.926 & 0.586 & 0.747 \\
\bottomrule
\end{tabular}}
\end{table}

This operating point first requires a survey that covers useful stopping regions without knowing future goals. The full-validation decomposition in Table~\ref{tab:coverage} shows 0.990 SR for target-view oracle coverage and 0.994 SR/0.992 SPL for the goal-independent 0.8\,m topology with evaluator-assigned semantics. Independent sampling therefore provides nearly complete success-region coverage without privileged target views. The remaining gap to real GPT-5.4 semantics localizes the main challenge to semantic grounding and STOP placement rather than map density; confidence intervals and failure counts appear in the supplement.

Coverage alone, however, does not solve retrieval. Even on privileged target-view poses, real Qwen semantics reach only 0.644/0.438 in Table~\ref{tab:coverage}. Supplementary Table~2 shows the progression behind that result: nearest-node retrieval gives 0.426/0.324, primary-label filtering gives 0.574/0.415, and confidence ranking reduces wrong selections from 549 to 331. Confidence clearly extracts value from fixed observations, but the remaining oracle gap motivates the proposed multi-view support and metric reachability.

\subsection{Metric Grounding and Fusion}

\begin{table}[t]
\centering
\caption{Cumulative full-validation component ablation on the same 1,000 episodes, 6,642-node topology, and 6,642 GPT-5.4 responses. SR/SPL are measured after at most $K$ destinations; DTG is shown for one-shot variants.}
\label{tab:main}
\resizebox{\columnwidth}{!}{\begin{tabular}{lccccrrr}
\toprule
Variant & Depth & Fusion & Recovery & $K$ & SR & SPL & DTG \\
\midrule
Camera-node baseline & \xmark & \xmark & \xmark & 1 & 0.835 & 0.560 & 1.039 \\
$+$ metric grounding & \cmark & \xmark & \xmark & 1 & 0.920 & \textbf{0.603} & 0.879 \\
$+$ source-aware fusion & \cmark & \cmark & \xmark & 1 & 0.926 & 0.586 & \textbf{0.747} \\
$+$ residual recovery & \cmark & \cmark & \cmark & 3 & \textbf{0.975} & 0.601 & -- \\
\bottomrule
\end{tabular}}
\end{table}

Having established sufficient coverage, we next ask which components convert observations into successful destinations. The cumulative ablation in Table~\ref{tab:main} assigns a distinct role to each stage. Metric grounding supplies the largest one-shot gain under fixed perception, changing 93 failures to successes and eight in reverse and raising SR/SPL by 0.085/0.042 (exact McNemar $p<10^{-18}$). Fusion compresses the primary index from 20,107 candidates to 1,329 supported hypotheses, lowers DTG from 0.879 to 0.747\,m, and reaches 0.926 one-shot SR. Residual recovery then raises completion to 0.975 SR/0.601 SPL. The gains therefore come from actionable geometry, source support, and candidate diversity rather than a change in VLM responses.

\begin{table}[t]
\centering
\caption{Same-response RGB-D and fusion ablations across scale, FoV, and backend. Each entry is SR/SPL; bold marks the best SR within each triplet. The full row uses 1,000 episodes/36 scenes; the remaining rows use 30 episodes/two scenes.}
\label{tab:controls}
\resizebox{\columnwidth}{!}{\begin{tabular}{llccc}
\toprule
Backend & FoV & Camera node & Raw RGB-D & Soft fusion \\
\midrule
GPT-5.4 (full) & $120^\circ$ & 0.835/0.560 & 0.920/0.603 & \textbf{0.926}/0.586 \\
GPT-5.4 (mini) & $90^\circ$ & 0.733/0.568 & 0.867/0.681 & \textbf{0.967}/0.665 \\
GPT-5.4 (mini) & $120^\circ$ & 0.933/0.704 & 0.933/0.713 & \textbf{1.000}/0.691 \\
\midrule
Qwen2.5-VL-3B (mini) & $90^\circ$ & 0.400/0.233 & 0.333/0.188 & \textbf{0.500}/0.218 \\
MiMo-v2.5 (mini) & $90^\circ$ & 0.200/0.129 & 0.833/0.614 & \textbf{0.900}/0.664 \\
MiMo-v2.5 (mini) & $120^\circ$ & 0.567/0.429 & 0.567/0.398 & \textbf{0.633}/0.367 \\
\bottomrule
\end{tabular}}
\end{table}

To test whether fusion is merely a GPT-specific prompting effect, we repeat the same-response comparison across backend and FoV. In all six triplets in Table~\ref{tab:controls}, fusion raises raw SR: 0.920 to 0.926 at full scale, 0.867 to 0.967 and 0.933 to 1.000 for GPT controls, 0.333 to 0.500 for Qwen, and 0.833 to 0.900 and 0.567 to 0.633 for MiMo. This consistency, together with the 20,107-to-1,329 primary-index reduction, supports fusion as a representation component rather than a model-specific artifact; uncertainty and paired discordance appear in supplementary Tables~5 and~7.

These controls also reveal when a wider camera is useful. The $120^\circ$ view exposes a low toilet missed by the standard camera, while fusion replaces an unsupported sofa singleton with a 22-source cluster, as illustrated in Figure~\ref{fig:cases}(a,b). Accordingly, the GPT-5.4 wide row in Table~\ref{tab:controls} removes both raw failures and reaches 1.000 SR, trading only 0.022 SPL for perfect completion. MiMo instead falls from 0.900 to 0.633 fused SR when widened. Wider FoV is therefore not a free gain; visibility, projection geometry, and VLM localization must be calibrated together.

\begin{figure}[t]
  \centering
  \includegraphics[width=\columnwidth]{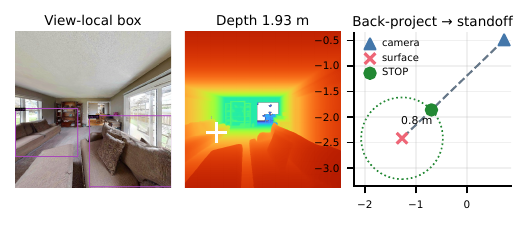}
  \caption{An audited sofa observation. A view-local VLM box indexes aligned depth; its median surface point is back-projected and shifted to a reachable 0.8\,m standoff. ObjectNav goals are not used in this construction.}
  \label{fig:grounding}
\end{figure}

The largest full-scale gain has a direct geometric explanation. In Figure~\ref{fig:grounding}, the VLM box selects a metric surface and depth back-projection moves the goal from the observation pose to a reachable 0.8\,m standoff. With identical detections, this operation raises SR from 0.835 to 0.920 in Table~\ref{tab:main}. The improvement therefore comes from closing the observation-to-STOP gap, not from changing semantic predictions.

\subsection{Density and Robustness Analysis}

\begin{figure}[t]
  \centering
  \includegraphics[width=\columnwidth]{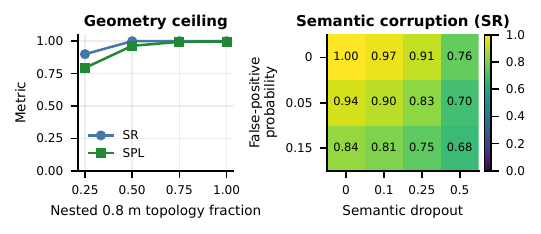}
  \caption{Geometric density saturates before semantic reliability on minival. Left: oracle semantics on nested goal-independent topologies. Right: mean SR over 20 controlled oracle-label corruptions with all nodes retained.}
  \label{fig:density}
\end{figure}

Once detections are grounded into reachable standoffs, adding topology nodes quickly stops helping. The left panel of Figure~\ref{fig:density} shows that 0.25/0.50/0.75/1.00 topology fractions use about 85/170/255/339 nodes yet reach oracle SR 0.900/1.000/1.000/1.000. Half the topology already covers every minival success region, making semantic localization the higher-value target beyond this density.

Semantic reliability behaves differently. In the right panel of Figure~\ref{fig:density}, 25\% and 50\% dropout reduce mean SR from 1.000 to 0.907 and 0.763, while 15\% false-positive probability reduces it to 0.840. The full 48-condition grid (28,800 evaluations) falls to 0.602 SR when 50\% node retention, 25\% dropout, and 15\% false positives are combined. The contrast is the useful observation: geometry saturates early, whereas modest semantic corruption remains damaging. These controlled perturbations do not model a particular VLM.

\begin{figure*}[t]
  \centering
  \includegraphics[width=0.96\textwidth]{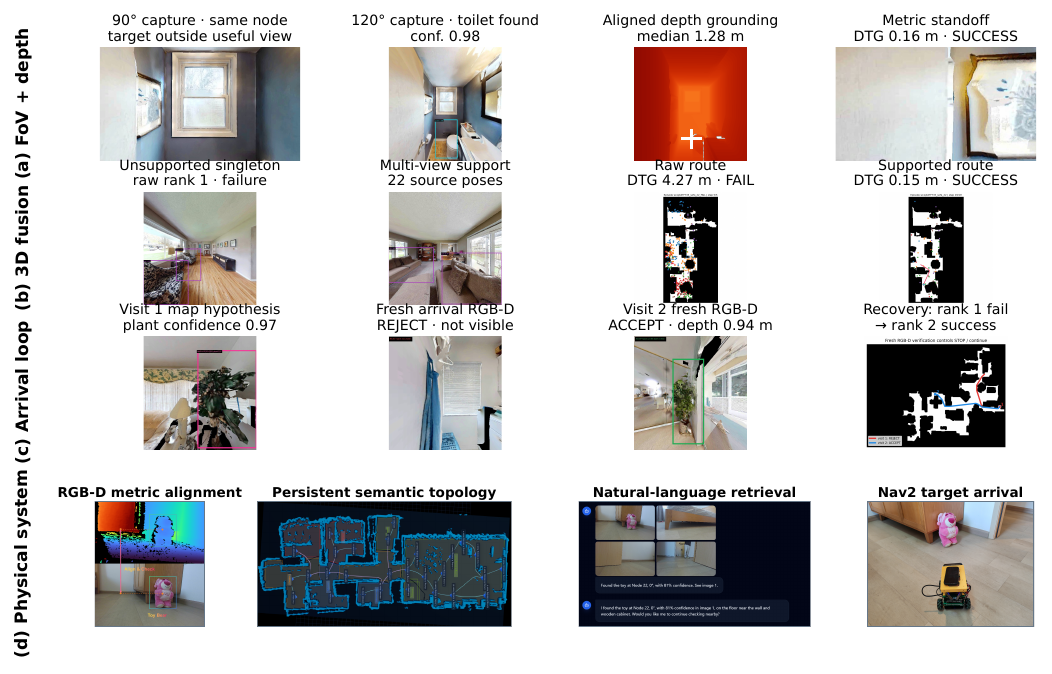}
  \caption{Mechanism and deployment audit. (a) Wide capture and depth ground a successful toilet standoff. (b) Source-aware fusion replaces an unsupported sofa singleton with a 22-view hypothesis. (c) Fresh RGB-D rejects a failed plant candidate and accepts rank 2. (d) The physical-system sequence shows onboard RGB-D alignment, the persistent semantic topology produced by the survey, natural-language retrieval of a toy at Node~22, and the robot's Nav2 arrival at the selected target. 
  }
  \label{fig:cases}
\end{figure*}

The mechanism-level audit closes the loop between these controls and execution. Fresh RGB-D rejects an incorrect rank-1 plant pose and accepts a spatially distinct rank-2 candidate in Figure~\ref{fig:cases}(c), the behavior quantified by Table~\ref{tab:topk}. The physical sequence in Figure~\ref{fig:cases}(d) traces the implemented robot pipeline from metric RGB-D observations and a surveyed semantic topology to language-conditioned retrieval and Nav2 target arrival.

\subsection{Sequential Fusion-Aware Recovery}

This failure mode motivates sequential fusion-aware recovery. We evaluate the first one, two, and three destinations in Table~\ref{tab:topk}; Success@$k$ asks whether the first $k$ visits contain a valid stopping pose, and SPL@$k$ accumulates executed legs through the first success. The protocol directly tests recovery from an incorrect primary hypothesis, while the deployed loop supplies the fresh RGB-D decision in Eq.~\ref{eq:arrival}; verifier calibration remains in the supplement.

\begin{table}[H]
\centering
\caption{Sequential recovery on full validation. Fused-only draws all visits from $\mathcal H^F$; ours draws one primary from $\mathcal H^F$ and 2\,m-separated backups from $\mathcal H^R$. $\Delta$SR uses the strict single fused target.}
\label{tab:topk}
\resizebox{\columnwidth}{!}{\begin{tabular}{lrrrrr}
\toprule
Candidate policy & $K$ & SR & $\Delta$SR & SPL & Successes \\
\midrule
Single fused target (no recovery) & 1 & 0.926 & -- & 0.586 & 926 \\
Fusion-aware primary only & 1 & 0.928 & +0.002 & 0.588 & 928 \\
Fused-only Top-3 control & 3 & 0.964 & +0.038 & 0.596 & 964 \\
Primary $+$ one diverse residual & 2 & 0.965 & +0.039 & 0.598 & 965 \\
Primary $+$ two diverse residuals & 3 & \textbf{0.975} & \textbf{+0.049} & \textbf{0.601} & \textbf{975} \\
\bottomrule
\end{tabular}}
\end{table}

The recovery ablation confirms that success comes from preserving useful alternatives, not simply revisiting more fused clusters. As Table~\ref{tab:topk} shows, a single target succeeds in 926 episodes; residual fallback adds two, and the first and second diverse residuals recover 37 and ten more. Final SR reaches 0.975 (+0.049) with 0.601 SPL, whereas fused-only Top-3 saturates at 0.964. The paired comparison yields eleven gains and no losses (nine sofa, two bed; exact McNemar $p<0.001$); matched raw and separation controls remain in the supplement.

\section{Physical-System Realization}

We deploy \sys{} on a Yahboom X3 mobile robot with ROS~2 Humble \cite{macenski2022ros2} and Nav2 \cite{macenski2020marathon2}, as shown in Figure~\ref{fig:cases}(d). A Gemini 336L supplies aligned RGB-D observations, an RPLIDAR S2 supplies laser scans, and \texttt{slam\_toolbox} \cite{macenski2021slamtoolbox} provides occupancy mapping and localization. The persistent manager retains $\mathcal G$ and interaction state across legs, linking natural-language category parsing, Eq.~\ref{eq:query-rank} planning, Nav2 execution, arrival verification, and candidate recovery. The illustrated query retrieves the toy at Node~22 before dispatching the selected target to Nav2; the corresponding real-robot run is provided separately as multimedia material.

\section{Discussion and Limitations}

Across protocols, goal-independent geometry reaches 0.994 SR, while metric grounding, fusion, and recovery convert that potential into 0.926 one-shot and 0.975 Top-3 SR (Tables~\ref{tab:coverage}, \ref{tab:main}, and \ref{tab:topk}). These gains establish topology coverage, object-centric metricization, source-aware evidence, and sequential recovery as a coherent route to reliable repeated ObjectNav.

\textbf{Limitations.} The full study covers one HM3D-v2 split, one hosted primary VLM, six categories, and static scenes. Continuous navmesh paths abstract control noise and map change, and relational dialogue lies beyond the category-query benchmark. Future work will extend the persistent-map and arrival interfaces to dynamic environments and broader real-camera evaluation.

\section{Conclusion}

\sys{} turns reusable RGB-D observations into reachable, fused metric-semantic goals. Grounding raises SR from 0.835 to 0.920, fusion reaches 0.926 one-shot SR, and residual recovery reaches 0.975 Top-3 SR (Tables~\ref{tab:main} and~\ref{tab:topk}). Together with the FoV/model controls and ROS~2/Nav2 realization, the result provides a reproducible foundation for high-reliability navigation over persistent semantic maps.

\ifdefined\SSTGCameraReady
\def\SSTGEmbeddedSupplement{}
\ifdefined\SSTGEmbeddedSupplement
\FloatBarrier
\appendix
\renewcommand{\thesection}{\arabic{section}}
\setcounter{section}{0}
\setcounter{table}{0}
\setcounter{figure}{0}
\setcounter{equation}{0}
\setcounter{secnumdepth}{2}
\section*{Supplementary Material}
\else
\documentclass[letterpaper]{article}
\usepackage[submission]{aaai2027}
\usepackage[hyphens]{url}
\usepackage{graphicx}
\urlstyle{rm}
\def\UrlFont{\rm}
\usepackage{natbib}
\usepackage{caption}
\usepackage{booktabs}
\usepackage{multirow}
\usepackage{amsmath}
\usepackage{amssymb}
\usepackage{pifont}
\frenchspacing

\pdfinfo{
/TemplateVersion (2027.1)
}
\setcounter{secnumdepth}{2}

\newcommand{\cmark}{\ding{51}}
\newcommand{\xmark}{\ding{55}}
\newcommand{\sys}{SSTG-Nav}
\setlength{\titlebox}{2.30in}

\title{SSTG-Nav: Metric-Grounded Spatial-Semantic Topological Graphs for Reusable Object Navigation\\Supplementary Material}
\author{Anonymous AAAI Submission}
\affiliations{Anonymous Institution}

\begin{document}
\maketitle
\fi

\ifdefined\SSTGEmbeddedSupplement
This appendix provides self-contained protocol definitions, complete quantitative results, qualitative analyses, and physical-system details that complement the main paper.
\else
This supplementary material provides detailed protocol definitions, additional quantitative results, qualitative analyses, physical-system details, and reproducibility information. File paths reported in the artifact-availability section are relative to the accompanying supplementary artifact. Navigation videos and the complete real-robot run are provided as accompanying multimedia.
\fi

\section{Protocol Definitions}

\subsection{Task and Evaluator}

The evaluation uses the HM3D-Semantics v0.2 ObjectNav episodes from the 2023 Habitat Navigation Challenge \cite{habitatchallenge2023,yadav2023hm3dsem}. The episodes use HM3D scenes \cite{ramakrishnan2021hm3d} and are rendered in Habitat-Sim 0.3.3 \cite{savva2019habitat}. For episode $e$, let $c_e$ be the requested category, $\mathbf{x}_e$ the official start, $\ell_e$ the official shortest path, and $\mathcal{Y}_{c_e}$ all official navigable goal viewpoints for the category in that scene. A selected candidate $\hat{\mathbf{s}}_e$ succeeds when its navmesh geodesic distance to $\mathcal{Y}_{c_e}$ is at most 1\,m. The evaluator records
\begin{align}
S_e &= \mathbf{1}\!\left[\min_{\mathbf{y}\in\mathcal{Y}_{c_e}}d_{\mathrm{geo}}(\hat{\mathbf{s}}_e,\mathbf{y})\leq1\right],\\
\mathrm{SPL}_e &= S_e\frac{\ell_e}{\max(\ell_e,p_e)},\\
\mathrm{DTG}_e &= \min_{\mathbf{y}\in\mathcal{Y}_{c_e}}d_{\mathrm{geo}}(\hat{\mathbf{s}}_e,\mathbf{y}).
\end{align}
If no semantic candidate exists, SPL is zero and DTG is the official start-to-goal distance. Episode keys concatenate scene, category, and original episode ID because HM3D reuses numeric IDs across categories. Path and goal-distance caches include candidate coordinates rather than map-local node IDs, and predicted categories never shortcut distance computation; only explicitly evaluator-created oracle nodes can do so. These implementation safeguards prevent semantic predictions from being treated as ground truth. SPL follows the standard embodied-navigation definition \cite{anderson2018evaluation}; the success interpretation follows ObjectNav recommendations \cite{batra2020objectnav}.

Main-paper Table~1 separates map supervision from post-hoc scoring. The independent real protocol is the primary result: episode starts, categories, object annotations, and valid goal viewpoints remain unavailable to topology construction, RGB-D capture, VLM inference, projection, fusion, and retrieval. The tables below expand that information boundary through construction counts, model-response completion statistics, and episode-level results.

\begin{table*}[t]
\centering
\caption{Complete published ObjectNav context retained outside the compact main table. HM3D-v1 and HM3D-v2 are shown separately; reusable-map methods amortize a survey over repeated queries. TF denotes no task-specific policy training, and CARe reports a custom MP3D success metric ($\dagger$).}
\label{tab:s-context}
\scriptsize
\setlength{\tabcolsep}{4.0pt}
\begin{tabular*}{\textwidth}{@{\extracolsep{\fill}}lllcrr}
\toprule
Method & Scene at start & Evaluation & Policy & SR & SPL \\
\midrule
\multicolumn{6}{l}{\emph{Unknown-scene HM3D-v1}}\\
VoroNav \cite{wu2024voronav} & unknown & HM3D-v1 & TF & 0.420 & 0.260 \\
VLFM \cite{yokoyama2024vlfm} & unknown & HM3D-v1 & TF & 0.525 & 0.304 \\
GAMap \cite{yuan2024gamap} & unknown & HM3D-v1 & TF & 0.531 & 0.260 \\
SG-Nav \cite{yin2024sgnav} & unknown & HM3D-v1 & TF & 0.540 & 0.249 \\
FBN \cite{zhang2025fbn} & unknown & HM3D-v1 & TF & 0.588 & 0.312 \\
ApexNav \cite{zhang2025apexnav} & unknown & HM3D-v1 & TF & 0.596 & 0.330 \\
TrajRAG \cite{wang2026trajrag} & unknown & HM3D-v1 & TF & 0.625 & 0.339 \\
CogNav \cite{cao2025cognav} & unknown & HM3D-v1 & TF & 0.725 & 0.262 \\
\midrule
\multicolumn{6}{l}{\emph{Unknown-scene HM3D-v2}}\\
SG-Nav \cite{yin2024sgnav,wang2026trajrag} & unknown & HM3D-v2 & TF & 0.496 & 0.255 \\
InstructNav \cite{long2024instructnav} & unknown & HM3D-v2 & TF & 0.580 & 0.209 \\
VLFM \cite{yokoyama2024vlfm,wang2026trajrag} & unknown & HM3D-v2 & TF & 0.636 & 0.325 \\
ApexNav \cite{zhang2025apexnav} & unknown & HM3D-v2 & TF & 0.762 & 0.380 \\
FOM-Nav \cite{chabal2025fomnav} & unknown & HM3D-v2 & learned & 0.758 & 0.479 \\
TrajRAG \cite{wang2026trajrag} & unknown & HM3D-v2 & TF & 0.781 & 0.402 \\
IntentNav \cite{cai2026intentnav} & unknown & HM3D-v2 & learned & 0.822 & 0.385 \\
ConsistNav \cite{wang2026consistnav} & unknown & HM3D-v2 & TF & 0.842 & 0.412 \\
\midrule
\multicolumn{6}{l}{\emph{Pre-explored reusable maps}}\\
CARe+VLMaps \cite{ko2024care} & pre-explored & custom MP3D & TF & 0.827$^\dagger$ & -- \\
\sys{} fused, single & pre-explored & HM3D-v2 & TF & 0.926 & 0.586 \\
\sys{} fusion-aware Top-3 & pre-explored & HM3D-v2 & TF & \textbf{0.975} & \textbf{0.601} \\
\bottomrule
\end{tabular*}
\end{table*}

Table~\ref{tab:s-context} preserves the broad comparison while allowing the main paper to foreground the most relevant HM3D-v2 block. Across 16 published unknown-scene rows spanning both HM3D versions, the reusable-map operating point remains visibly distinct; the controlled internal tables below then isolate where \sys{} obtains its gains.

\subsection{Protocol A: Target-View Coverage}

For every annotated object instance with available viewpoints, the mapper chooses one navigable high-IoU official viewpoint and captures four RGB views at yaw offsets $0^\circ$, $-45^\circ$, $45^\circ$, and $180^\circ$. Across the 36 validation scenes, this creates 1,168 nodes and 6,446 edges. Because official goal viewpoints determine capture locations, this protocol measures semantic retrieval and planning under controlled coverage. It is neither goal-independent nor deployable.

\begin{table}[t]
\centering
\caption{Full HM3D-v2 validation under target-view coverage (1,000 episodes, 36 scenes).}
\label{tab:s-full}
\resizebox{\columnwidth}{!}{\begin{tabular}{lrrrrl}
\toprule
Variant & SR & SR 95\% CI & SPL & SPL 95\% CI & Failures \\
\midrule
Oracle labels & 0.990 & [0.982,0.995] & 0.802 & [0.791,0.812] & 10 missing \\
Qwen all, nearest & 0.426 & [0.396,0.457] & 0.324 & [0.299,0.349] & 549 wrong, 25 missing \\
Qwen primary & 0.574 & [0.543,0.604] & 0.415 & [0.389,0.439] & 265 wrong, 161 missing \\
Qwen all, confidence & 0.644 & [0.614,0.673] & 0.438 & [0.415,0.461] & 331 wrong, 25 missing \\
\bottomrule
\end{tabular}}
\end{table}

Table~\ref{tab:s-full} shows that target-view coverage is not exactly one. Ten episodes have no retained coverage node: four TV/monitor episodes in scene \texttt{BAbdmeyTvMZ}, five plant episodes in \texttt{q5QZSEeHe5g}, and one chair episode in the same scene. These cases are retained. Qwen inference completes on 1,167 of 1,168 node images; the unsuccessful response is also retained as a mapping failure. The 0.346 SR gap between oracle and Qwen under the same privileged poses isolates semantic retrieval error from coverage.

The per-category target-view audit is incorporated into Table~\ref{tab:s-category} alongside the full goal-independent results. Under confidence-ranked Qwen semantics, TV/monitor is lowest at 0.430 SR whereas bed reaches 0.818. These values describe semantic retrieval under controlled coverage and are not used as class weights or tuning signals.

\subsection{Protocol B: Goal-Independent Topology}

For each scene, Habitat's pathfinder draws 12,000 navigable points using seed 20260719 plus the deterministic scene index. Greedy farthest-point selection starts at the pool point nearest its centroid and repeatedly adds the point farthest from the selected set until every pool point is within 0.8\,m. Up to eight Euclidean-nearest nodes are considered for edges. An edge is kept only when the Euclidean distance is at most 2.4\,m, a navmesh route exists, and its geodesic length is at most 3.84\,m. The mapper uses the split only to enumerate scene assets; no episode start, category, object annotation, or goal viewpoint is consulted by sampling, capture, inference, or fusion.

\begin{table}[t]
\centering
\caption{Independent topology construction. The full split has 6,642 nodes over 36 scenes; per-scene counts range from 74 to 399.}
\label{tab:s-topology}
\resizebox{\columnwidth}{!}{\begin{tabular}{lrrr}
\toprule
Split / scene & Nodes & Edges & Empirical cover radius \\
\midrule
Full validation & 6,642 & 21,845 & $\leq0.800$\,m \\
\midrule
Minival: \texttt{TEEsavR23oF} & 132 & 415 & $\leq0.8$\,m \\
Minival: \texttt{wcojb4TFT35} & 207 & 691 & $\leq0.8$\,m \\
Minival total & 339 & 1,106 & -- \\
\bottomrule
\end{tabular}}
\end{table}

Table~\ref{tab:s-topology} verifies that the independent construction covers both compact and large scenes with one radius rule rather than a fixed node count. Oracle semantics are assigned only after mapping by measuring each node against the official category goal sets. This evaluator-side labeling produces a geometric coverage ceiling of 0.994 SR/0.992 SPL on all 1,000 validation episodes (994 successes; SR 95\% CI [0.987,0.997]). Real-semantic variants never receive those labels.

\section{Metric Semantic Mapping Details}

\subsection{RGB-D Capture and Projection}

Every independent node stores four cardinal views at yaw offsets $0^\circ$, $90^\circ$, $180^\circ$, and $270^\circ$. The standard camera is $640\times360$ with $90^\circ$ horizontal FoV; the wide camera is $640\times640$ with $120^\circ$ horizontal FoV. RGB and depth share a pose 1.25\,m above the agent base. Habitat depth is camera-forward $Z$ depth. For image width $W$, height $H$, and horizontal FoV $\phi_h$, vertical FoV is
\begin{equation}
\phi_v=2\tan^{-1}\!\left(\frac{H}{W}\tan\frac{\phi_h}{2}\right).
\end{equation}
The standard camera consequently has an approximately $59^\circ$ vertical FoV, whereas the square wide camera has $120^\circ$ in both directions.

The VLM returns category, view index, normalized box, and confidence. We use the box center and the median of valid depth values in a $7\times7$ patch, accepting 0.2--6.0\,m. Let $(u,v)$ be normalized center coordinates and $z$ depth. Camera-frame coordinates are
\begin{align}
x_c&=2(u-0.5)z\tan(\phi_h/2),\\
y_c&=-2(v-0.5)z\tan(\phi_v/2),\qquad z_c=-z.
\end{align}
The stored node rotation and view yaw transform this point to world coordinates. A desired 0.8\,m standoff is placed toward the source camera in the horizontal plane, assigned the source base height, snapped to the navmesh, and rejected when unreachable from the observing node.

\subsection{Fusion Parameters and Decision Rules}

Candidates are first partitioned by category. Two candidates are connected when their object estimates are within 1.2\,m horizontally and 1.0\,m vertically, and their snapped STOP poses are mutually reachable within 3.0\,m geodesic distance. Connected components define clusters. Within cluster $C$, duplicate detections from one source node contribute only their maximum confidence $q_i$, producing
\begin{equation}
Q(C)=1-\prod_{i\in U(C)}(1-\min(0.99,q_i)).
\end{equation}
For the full-validation map, clusters with at least two unique source nodes are retained and a singleton is retained when its confidence is at least 0.92. The minival representation/model controls retain every valid cluster (minimum support one), because applying the full-map filter to only two scenes would confound representation with category deletion. In both cases the highest-confidence member supplies the representative STOP pose; clustering geometry and noisy-OR scoring are unchanged. The main paper therefore compares representations within, not across, each scale/backend triplet.

The reachability condition is necessary. Euclidean proximity alone can merge observations across walls or between disconnected navmesh islands. Likewise, repeated boxes from one frame are not independent confirmation and must not increase support.

\subsection{VLM Prompt and Parser}

Both hosted backends receive four separate RGB images in one node-level request. The effective instruction is:
\begin{quote}\small
Inspect the four indexed views from one robot pose. Detect every visible instance of chair, bed, plant, toilet, TV/monitor, and sofa. For each detection return JSON with \texttt{view\_index}, canonical \texttt{category}, normalized \texttt{bbox\_norm=[x1,y1,x2,y2]}, and \texttt{confidence}. Boxes must be local to the named view; do not use panorama coordinates or infer an object that is not visibly grounded.
\end{quote}
The parser accepts coordinates in $[0,1]$ or $[0,1000]$, converts the latter, clamps bounds, rejects invalid boxes and categories, and records the raw response unchanged. GPT-5.4 \cite{openai2026gpt54} and MiMo-v2.5 \cite{mimov25} use the same parser version, \texttt{bbox-norm-or-1000-v2}, so model comparisons do not confound response scaling.

Local Qwen2.5-VL-3B \cite{bai2025qwen25vl} uses the same four $90^\circ$ RGB-D views in a $2\!\times\!2$ grounding panorama and returns an absolute box confined to one tile. Valid empty lists are retained. A subset of panoramas produces degenerate non-JSON text under deterministic decoding; those failed nodes are repaired by applying a simpler, identical category-and-box prompt independently to the four original views. If a view remains token-degenerate, the same RGB is re-encoded at a nearby patch grid for one final retry. The cache records \texttt{repair\_mode=four\_view\_local\_grounding}. Camera-node, raw, and fused Qwen variants all consume the resulting immutable detection cache, so the repair changes inference packaging but does not confound the representation comparison.

\begin{table*}[t]
\centering
\caption{Completed API and candidate-generation records for the independent topology. Each request contains four RGB views. Token totals are provider-reported and are not directly comparable across providers.}
\label{tab:s-api}
\resizebox{\textwidth}{!}{\begin{tabular}{llrrrrrr}
\toprule
Backend & Sensor & Requests & Completed & Raw detections & Valid depth candidates & Fused candidates & Tokens \\
\midrule
GPT-5.4 full & $120^\circ$, $640\times640$ & 6,642 & 6,642 & 21,579 & 20,107 & 1,329 & 44,499,374 \\
\midrule
GPT-5.4 & $90^\circ$, $640\times360$ & 339 & 339 & 746 & 674 & 129 & 2,031,743 \\
GPT-5.4 & $120^\circ$, $640\times640$ & 339 & 339 & 971 & 922 & 130 & 2,305,616 \\
MiMo-v2.5 & $90^\circ$, $640\times360$ & 339 & 339 & 794 & 735 & 125 & 461,903 \\
MiMo-v2.5 & $120^\circ$, $640\times640$ & 339 & 339 & 1,022 & 969 & 108 & 715,700 \\
Qwen2.5-VL-3B & $90^\circ$, $640\times360$ & 339 & 339 & 1,064 & 959 & 315 & local \\
\bottomrule
\end{tabular}}
\end{table*}

Table~\ref{tab:s-api} establishes completion and representation scale: full fusion compresses 20,107 valid depth candidates to 1,329, while every node-level request completes. The complete independent run uses model string \texttt{gpt-5.4} through a hosted chat-compatible endpoint.
\ifdefined\SSTGEmbeddedSupplement
\else
The configured \texttt{gpt-5.5} Responses-compatible route returned HTTP 502 on a minimal request during this run; it is not silently renamed or mixed with the completed model. A separate target-view minival pilot had previously completed with \texttt{gpt-5.5} and is reported only as a semantic-isolation result below.
\fi

\section{Complete Results}

\subsection{Independent RGB-D Variants}

\begin{table*}[t]
\centering
\caption{Independent-topology results. ``Full'' rows use 1,000 episodes/36 scenes; remaining rows use 30 episodes/two scenes. SR intervals are Wilson; SPL intervals use 10,000 episode-level bootstrap resamples.}
\label{tab:s-rgbd}
\begin{tabular}{lllrrrrrr}
\toprule
Backend & Representation & Sensor & SR & SR 95\% CI & SPL & SPL 95\% CI & DTG \\
\midrule
Oracle & evaluator geometry & full, $0.8$\,m & 0.994 & [0.987,0.997] & 0.992 & [0.987,0.996] & 0.622 \\
GPT-5.4 full & camera node & $120^\circ$ & 0.835 & [0.811,0.857] & 0.560 & [0.539,0.582] & 1.039 \\
GPT-5.4 full & raw RGB-D & $120^\circ$ & 0.920 & [0.902,0.935] & 0.603 & [0.584,0.621] & 0.879 \\
GPT-5.4 full & soft fusion & $120^\circ$ & 0.926 & [0.908,0.941] & 0.586 & [0.568,0.604] & 0.747 \\
\midrule
Qwen2.5-VL-3B & category panorama & $90^\circ$ & 0.400 & [0.246,0.577] & 0.344 & [0.195,0.501] & 2.812 \\
Qwen2.5-VL-3B & localized camera & $90^\circ$ & 0.400 & [0.246,0.577] & 0.233 & [0.125,0.347] & 5.412 \\
Qwen2.5-VL-3B & raw RGB-D & $90^\circ$ & 0.333 & [0.192,0.512] & 0.188 & [0.091,0.294] & 5.970 \\
Qwen2.5-VL-3B & soft fusion & $90^\circ$ & 0.500 & [0.332,0.668] & 0.218 & [0.128,0.319] & 5.791 \\
GPT-5.4 & camera node & $90^\circ$ & 0.733 & [0.556,0.858] & 0.568 & [0.422,0.706] & 1.436 \\
GPT-5.4 & raw RGB-D & $90^\circ$ & 0.867 & [0.703,0.947] & 0.681 & [0.569,0.779] & 0.560 \\
GPT-5.4 & soft fusion & $90^\circ$ & 0.967 & [0.833,0.994] & 0.665 & [0.580,0.743] & 0.375 \\
GPT-5.4 & camera node & $120^\circ$ & 0.933 & [0.787,0.982] & 0.704 & [0.606,0.796] & 0.424 \\
GPT-5.4 & raw RGB-D & $120^\circ$ & 0.933 & [0.787,0.982] & 0.713 & [0.616,0.801] & 0.440 \\
GPT-5.4 & soft fusion & $120^\circ$ & 1.000 & [0.886,1.000] & 0.691 & [0.609,0.771] & 0.136 \\
MiMo-v2.5 & camera node & $90^\circ$ & 0.200 & [0.095,0.373] & 0.129 & [0.039,0.232] & 2.418 \\
MiMo-v2.5 & raw RGB-D & $90^\circ$ & 0.833 & [0.664,0.927] & 0.614 & [0.482,0.736] & 1.414 \\
MiMo-v2.5 & soft fusion & $90^\circ$ & 0.900 & [0.744,0.965] & 0.664 & [0.550,0.765] & 1.011 \\
MiMo-v2.5 & camera node & $120^\circ$ & 0.567 & [0.392,0.726] & 0.429 & [0.275,0.581] & 5.979 \\
MiMo-v2.5 & raw RGB-D & $120^\circ$ & 0.567 & [0.392,0.726] & 0.398 & [0.257,0.539] & 5.981 \\
MiMo-v2.5 & soft fusion & $120^\circ$ & 0.633 & [0.455,0.781] & 0.367 & [0.245,0.488] & 3.840 \\
\bottomrule
\end{tabular}
\end{table*}

Table~\ref{tab:s-rgbd} gives the complete uncertainty record behind the compact main-paper table. The full rows use the same 6,642-node goal-independent topology; minival rows use its separately sampled 339-node control graph. Within each localized backend/FoV group, camera-node, raw, and fused variants reuse identical cached detections. The category-only Qwen panorama row is retained as a separate legacy coverage result and is not part of the localized triplet. Qwen camera-to-raw has 0/2 gains/losses, raw-to-fused has 8/3, and camera-to-fused has 6/3; inaccurate single boxes and useful repeated evidence therefore coexist, and fusion is not monotonic. Full GPT fusion changes 28 raw failures to successes and 22 successes to failures, so its net SR gain is modest even though it reduces candidate count by 93.4\%. GPT-5.4 wide raw fails two sofa episodes and fusion succeeds on all episodes; MiMo failures show that wide input and depth do not compensate for a weaker localization response distribution.

\begin{table*}[t]
\centering
\caption{Consolidated full-validation per-category audit. Entries are SR/SPL over the same 1,000 episodes. Target-view Qwen isolates semantic retrieval under controlled coverage; the three goal-independent GPT-5.4 columns isolate spatial representation.}
\label{tab:s-category}
\begin{tabular}{lrrrr}
\toprule
& Target-view semantic isolation & \multicolumn{3}{c}{Goal-independent metric-semantic mapping} \\
\cmidrule(lr){2-2}\cmidrule(lr){3-5}
Category ($n$) & Qwen confidence & Camera node & Raw RGB-D & Soft fusion \\
\midrule
Chair (195) & 0.708/0.454 & 0.841/0.415 & 0.974/0.488 & 0.995/0.498 \\
Bed (165) & 0.818/0.550 & 0.891/0.673 & 0.988/0.716 & 0.976/0.631 \\
Plant (152) & 0.724/0.462 & 0.862/0.473 & 0.868/0.481 & 0.941/0.532 \\
Toilet (166) & 0.633/0.440 & 0.934/0.771 & 0.940/0.746 & 0.940/0.673 \\
TV/monitor (135) & 0.430/0.356 & 0.644/0.455 & 0.800/0.578 & 0.778/0.571 \\
Sofa (187) & 0.524/0.361 & 0.807/0.572 & 0.914/0.611 & 0.893/0.613 \\
\bottomrule
\end{tabular}
\end{table*}

Table~\ref{tab:s-category} retains the complete full-scale category workload without mixing it with two-scene counts. Metric grounding improves most categories; fusion raises plant SR from 0.868 to 0.941 and chair SR to 0.995, while bed and sofa expose the remaining ranking-efficiency tradeoff.
\ifdefined\SSTGEmbeddedSupplement
\else
The accompanying artifact includes the corresponding 30-episode category breakdown; its one toilet and five plant episodes are too sparse for a second paper table.
\fi

\subsection{Paired Analyses}

We use exact two-sided McNemar tests for paired success outcomes and a 10,000-resample paired episode bootstrap for SPL differences. These analyses were added after the primary summaries and do not select a model or hyperparameter.

\begin{table}[t]
\centering
\caption{Paired changes. ``Gain/loss'' counts success discordances from the first variant to the second.}
\label{tab:s-paired}
\resizebox{\columnwidth}{!}{\begin{tabular}{lrrrr}
\toprule
Comparison & Gain/loss & $p$ & $\Delta$SPL & 95\% CI \\
\midrule
Full camera $\to$ raw & 93/8 & $1.74\!\times\!10^{-19}$ & 0.042 & [0.028,0.057] \\
Full raw $\to$ fused & 28/22 & 0.480 & $-0.017$ & [$-0.032$,$-0.002$] \\
Full camera $\to$ fused & 113/22 & $6.09\!\times\!10^{-16}$ & 0.026 & [0.005,0.047] \\
\midrule
GPT $90^\circ$ raw $\to$ fused & 3/0 & 0.250 & -0.016 & [-0.074,0.047] \\
GPT $90^\circ$ camera $\to$ raw & 7/3 & 0.344 & 0.113 & [-0.014,0.243] \\
GPT $90^\circ$ camera $\to$ fused & 7/0 & 0.0156 & 0.097 & [-0.014,0.211] \\
GPT fused $90^\circ\to120^\circ$ & 1/0 & 1.000 & 0.026 & [-0.038,0.111] \\
Qwen raw $\to$ fused & 8/3 & 0.227 & 0.030 & [-0.103,0.167] \\
MiMo fused $90^\circ\to120^\circ$ & 0/8 & 0.0078 & -0.297 & [-0.439,-0.157] \\
MiMo $90^\circ$ camera $\to$ fused & 21/0 & $<0.0001$ & 0.535 & [0.402,0.662] \\
\bottomrule
\end{tabular}}
\end{table}

Table~\ref{tab:s-paired} shows that the same-response camera-node contrasts are not merely a GPT-versus-Qwen model effect. On full validation, depth grounding supplies the dominant gain; fusion compresses the map and slightly raises SR, but reduces SPL. GPT-5.4's mini $90^\circ$ camera-to-fusion success change is detectable, Qwen's fused SR advantage comes from 8/3 raw-to-fused discordances, and MiMo exposes both a strong standard-FoV metricization gain and significant wide-FoV degradation.
\ifdefined\SSTGEmbeddedSupplement
\else
The accompanying artifact includes the complete paired matrices and the omitted redundant camera/fusion contrasts.
\fi
No multiple-testing correction is applied, so these values are targeted diagnostics rather than a confirmatory family of tests.

\subsection{Topology Density}

The topology is nested: the first quarter, half, and three quarters of the farthest-point order are strict subsets of the full map. Official goals are loaded after mapping to label candidates for this geometric ceiling only.

\begin{table}[t]
\centering
\caption{Independent topology density ceiling on minival.}
\label{tab:s-density}
\resizebox{\columnwidth}{!}{\begin{tabular}{rrrrr}
\toprule
Node fraction & Approx. nodes & SR & SPL & DTG \\
\midrule
0.25 & 85 & 0.900 & 0.794 & 0.784 \\
0.50 & 170 & 1.000 & 0.964 & 0.376 \\
0.75 & 255 & 1.000 & 0.995 & 0.508 \\
1.00 & 339 & 1.000 & 0.997 & 0.555 \\
\bottomrule
\end{tabular}}
\end{table}

Table~\ref{tab:s-density} shows that oracle SR saturates at half density, whereas SPL continues to improve as candidates approach shorter routes. DTG is not monotonic after SR saturates because any point within the 1\,m goal set counts as successful, and added candidates can be closer to the episode start without being the closest point to the target. SPL is computed with the official $\max(\ell,p)$ denominator and remains bounded by one.

\subsection{Sequential Candidates}

Table~\ref{tab:s-topk} deliberately uses official goal distance after every visit. It measures candidate-list potential and remains separate from the autonomous verifier evaluated next.

\begin{table*}[t]
\centering
\caption{Sequential candidate and separation sensitivity. A positive separation greedily removes candidates near earlier hypotheses. The full-scale policy block exposes both the matched 2\,m comparison and nearby sensitivity controls.}
\label{tab:s-topk}
\begin{tabular}{llrrrrrr}
\toprule
Map & Separation & Success@1 & Success@2 & Success@3 & SPL@1 & SPL@2 & SPL@3 \\
\midrule
Qwen panorama & none & 0.400 & 0.500 & 0.500 & 0.344 & 0.396 & 0.396 \\
Qwen panorama & 5.0\,m & 0.400 & 0.567 & 0.700 & 0.344 & 0.392 & 0.425 \\
GPT-5.4 raw $120^\circ$ & none & 0.933 & 1.000 & 1.000 & 0.713 & 0.745 & 0.745 \\
\midrule
GPT-5.4 full raw & none & 0.920 & 0.955 & 0.957 & 0.603 & 0.610 & 0.611 \\
GPT-5.4 full raw & 2\,m & 0.920 & 0.963 & \textbf{0.975} & 0.603 & 0.612 & \textbf{0.616} \\
GPT-5.4 full raw & 3\,m & 0.920 & 0.965 & 0.975 & 0.603 & 0.614 & 0.616 \\
GPT-5.4 full fused & none & 0.926 & 0.952 & 0.964 & 0.586 & 0.594 & 0.596 \\
GPT-5.4 full fused & 2\,m & 0.926 & 0.952 & 0.964 & 0.586 & 0.594 & 0.596 \\
GPT-5.4 full fused & 3\,m & 0.926 & 0.950 & 0.962 & 0.586 & 0.593 & 0.596 \\
GPT-5.4 fusion-aware residual & 2\,m & \textbf{0.928} & \textbf{0.965} & \textbf{0.975} & 0.588 & 0.598 & 0.601 \\
\bottomrule
\end{tabular}
\end{table*}

\ifdefined\SSTGEmbeddedSupplement
The auxiliary Qwen sweep covers five separation values; its endpoints and best S@3 are shown here.
\else
The auxiliary Qwen sweep covers five separation values; its endpoints and best S@3 are shown here, and the complete sweep is included in the accompanying artifact.
\fi
On mini, GPT-5.4's two raw errors are recovered by the second candidate. At full scale, raw spatial diversity reaches the same 0.975 three-visit ceiling but starts at 0.920, whereas fusion-aware recovery reaches 0.928/0.965 after one/two visits and preserves 0.975 at visit three. The fused-only row draws all three destinations from the 1,329 cluster representatives. The fusion-aware row draws its first destination from that representative index, then selects 2\,m-separated backups from the 20,107 pre-fusion metric standoffs. This linked residual bank retains alternative stopping geometry suppressed by one-representative-per-cluster fusion and also provides a fallback when no cluster is retained. It yields eleven paired gains and no losses over fused-only Top-3 (nine sofa and two bed episodes). The official evaluator supplies the metric stop signal in every Top-$K$ row.

\subsection{Fresh-Arrival RGB-D/VLM Verification}

The deployable loop never reads the goal set. For each unique ranked candidate it captures four fresh square $120^\circ$ RGB-D views after arrival. View zero faces the stored 3D object estimate; the remaining views rotate by $90^\circ$. GPT-5.4 receives these four images and a boxed mapping-time reference, then returns target visibility, stopping-side validity, confidence, an arrival-view index, and a normalized box. The central half of that box indexes aligned depth. Strict acceptance requires visible target, valid stopping side, confidence at least 0.75, and median depth in $[0.25,2.5]$\,m. On rejection, the next geodesic starts at the current candidate.

The 1,000 raw-list episodes reference 656 unique candidates, producing 2,624 fresh RGB-D views and 656/656 completed verifier requests. They consume 4,731,659 tokens with mean/median/95th-percentile latency 20.52/19.45/32.31 seconds per unique candidate. Goal annotations are absent from capture, prompt construction, inference, and STOP/continue; they are read only afterward to form the confusion matrix and ObjectNav metrics.

\begin{table*}[t]
\centering
\caption{Autonomous full-validation arrival verification. S@$k$ and SPL@$k$ execute at most $k$ candidates. P/R compares verifier decisions with the post-hoc official success test over attempted visits. The RGB-D-only rule reparses the same cached responses but ignores the VLM stopping-side flag.}
\label{tab:s-arrival}
\resizebox{\textwidth}{!}{\begin{tabular}{llrrrrrrrrr}
\toprule
Candidates & Decision rule & S@1 & S@2 & S@3 & SPL@1 & SPL@2 & SPL@3 & P & R & Attempts \\
\midrule
Raw, 3\,m & strict dual geometry & 0.854 & 0.909 & 0.912 & 0.560 & 0.573 & 0.573 & 0.964 & 0.915 & 1.110 \\
Raw, 3\,m & category + RGB-D range & 0.865 & 0.903 & 0.905 & 0.569 & 0.578 & 0.578 & 0.956 & 0.924 & 1.091 \\
Fused & strict dual geometry & 0.821 & 0.901 & 0.901 & 0.526 & 0.546 & 0.546 & 0.947 & 0.891 & 1.087 \\
\bottomrule
\end{tabular}}
\end{table*}

Table~\ref{tab:s-arrival} calibrates the arrival classifier separately from candidate quality. Strict raw verification ends 912 episodes successfully, with candidate ranks 1/2/3 accounting for 888/55/3 accepts. The collapsed-representative fused ablation uses 572 unique candidates and 2,288 arrival views; exact observations reuse 314 raw-list responses and 258 new requests. This ablation exposed the implementation issue corrected by the final hierarchy: fusion now controls the primary hypothesis while residual metric standoffs remain available for Top-3 recovery.

\subsection{Coverage-Controlled Corruption}

The target-view oracle minival topology is perturbed across retained-node fractions $\{1,0.75,0.5,0.25\}$, semantic dropout probabilities $\{0,0.1,0.25,0.5\}$, and false-positive probabilities $\{0,0.05,0.15\}$. Every condition uses 20 seeds, giving 960 map perturbations and 28,800 episode evaluations. A false positive adds a randomly selected incorrect benchmark category to a retained node; dropout independently removes a correct label.

\begin{table*}[t]
\centering
\caption{Complete corruption grid: mean SR over 20 seeds. Each three-column group is false-positive probability 0/0.05/0.15.}
\label{tab:s-stress}
\resizebox{\textwidth}{!}{\begin{tabular}{crrrrrrrrrrrr}
\toprule
\multirow{2}{*}{Nodes} & \multicolumn{3}{c}{Dropout 0} & \multicolumn{3}{c}{Dropout 0.10} & \multicolumn{3}{c}{Dropout 0.25} & \multicolumn{3}{c}{Dropout 0.50} \\
 & FP 0 & FP 0.05 & FP 0.15 & FP 0 & FP 0.05 & FP 0.15 & FP 0 & FP 0.05 & FP 0.15 & FP 0 & FP 0.05 & FP 0.15 \\
\midrule
1.00 & 1.000 & 0.937 & 0.840 & 0.970 & 0.898 & 0.805 & 0.907 & 0.828 & 0.752 & 0.763 & 0.700 & 0.680 \\
0.75 & 0.907 & 0.893 & 0.768 & 0.885 & 0.853 & 0.730 & 0.833 & 0.807 & 0.713 & 0.690 & 0.653 & 0.642 \\
0.50 & 0.768 & 0.715 & 0.682 & 0.738 & 0.695 & 0.653 & 0.678 & 0.647 & 0.602 & 0.570 & 0.545 & 0.520 \\
0.25 & 0.530 & 0.497 & 0.497 & 0.488 & 0.455 & 0.458 & 0.430 & 0.407 & 0.432 & 0.347 & 0.333 & 0.363 \\
\bottomrule
\end{tabular}}
\end{table*}

Table~\ref{tab:s-stress} shows that dropout and false positives compound rather than substitute for one another: at 50\% retained nodes, 0.25 dropout and 0.15 false positives reduce mean SR to 0.602.
\ifdefined\SSTGEmbeddedSupplement
\else
The accompanying artifact includes the individual standard deviations and SPL values.
\fi
This experiment is a sensitivity analysis on a privileged target-view map. It does not estimate the error distribution of Qwen, GPT-5.4, or MiMo.

\ifdefined\SSTGEmbeddedSupplement
\else
\subsection{GPT-5.5 Semantic-Isolation Pilot}

The target-view minival map has 58 object-instance nodes and is independent of the later RGB-D run. All 58 \texttt{gpt-5.5} requests completed. Confidence ranking achieves 0.967 SR/0.639 SPL; nearest-node retrieval achieves 0.367/0.271 and primary-label retrieval achieves 0.567/0.428. The sole confidence-ranked failure is a missing toilet label. Because mapping poses are goal-derived, this is evidence about semantic retrieval only.

A full-validation \texttt{gpt-5.5} cache is incomplete: 642 of 1,168 requests completed, 460 ended with HTTP 502, and 66 with HTTP 429 after retry limits. We do not report a full-validation score from that subset. The complete full-validation real-semantics result instead uses local Qwen, whose inference completion is 1,167/1,168.
\fi

\section{Qualitative and Visual Audit}

\begin{figure*}[t]
  \centering
  \includegraphics[width=0.90\textwidth]{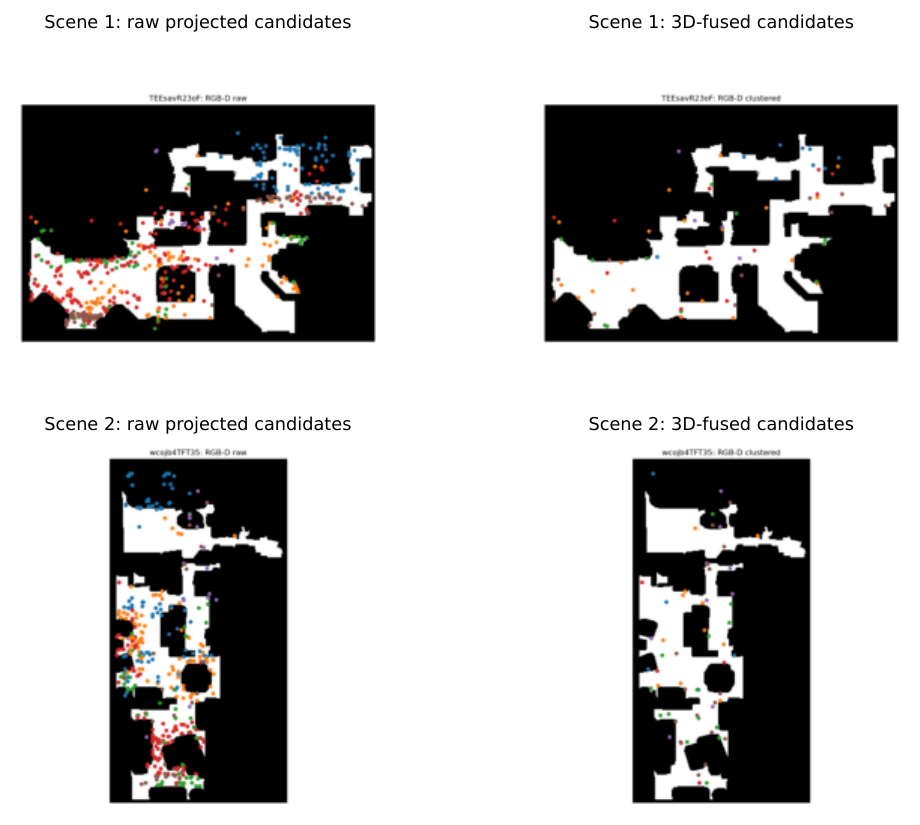}
  \caption{Actual semantic maps for both minival scenes. Each row uses a common Habitat top-down frame. Raw panels show every valid depth-projected candidate; fused panels show the candidates remaining after 3D and reachability clustering. Official goal regions are absent.}
  \label{fig:s-topology}
\end{figure*}

\begin{figure*}[t]
  \centering
  \includegraphics[width=0.98\textwidth]{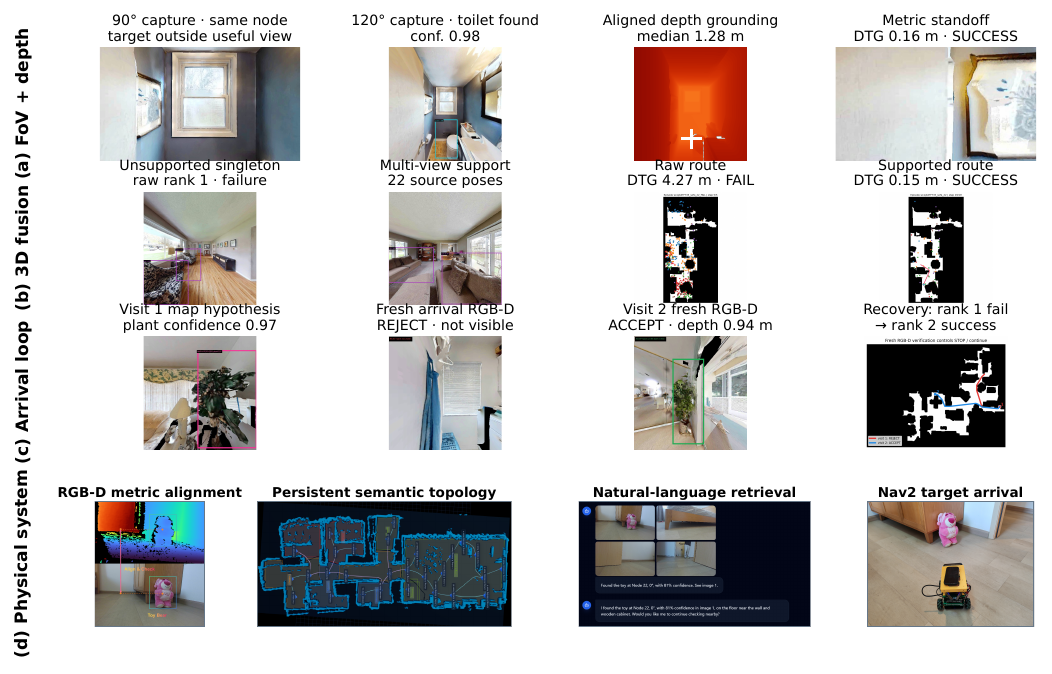}
  \caption{Source-traceable FoV, fusion, closed-loop arrival, and physical-system audit. Panels (a)--(c) link to their original RGB/depth file, pose, VLM box, candidate, and evaluated episode. The third row is a full-validation plant case: rank 1 is truly outside the goal set and rejected from fresh RGB-D; rank 2 is accepted at 0.94\,m measured depth and succeeds. Panel (d) records the physical pipeline: RGB-D alignment, the surveyed semantic topology, language-conditioned toy retrieval, and Nav2 target arrival.}
  \label{fig:s-cases}
\end{figure*}

\begin{figure*}[t]
  \centering
  \includegraphics[width=0.94\textwidth]{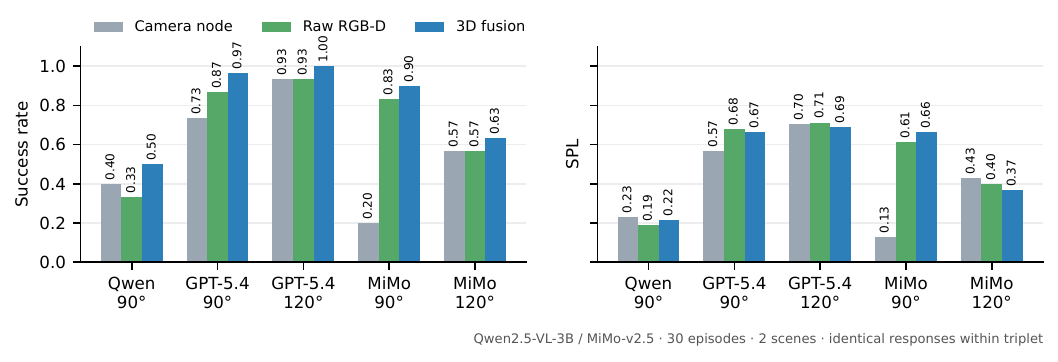}
  \caption{Same-response representation and model ablations on independent-topology minival. Each backend/FoV triplet reuses exact cached VLM responses and changes only whether category evidence is attached to the camera, back-projected with depth, or fused in 3D.}
  \label{fig:s-ablation}
\end{figure*}

\begin{figure*}[t]
  \centering
  \includegraphics[width=0.96\textwidth]{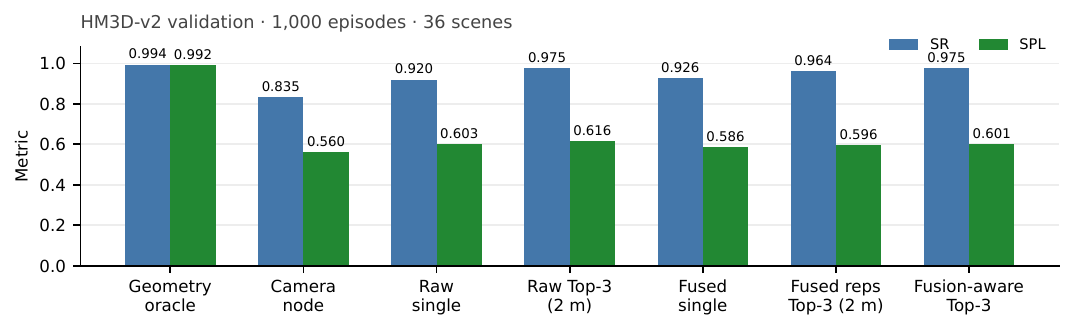}
  \caption{Full-validation independent-topology results. Geometry is the evaluator-labeled ceiling; camera, raw, and fused variants reuse identical GPT-5.4 mapping responses. Under matched 2\,m diversity, raw and fusion-aware Top-3 both reach 0.975 S@3; Table~\ref{tab:s-topk} resolves the stronger early recovery of the fusion-aware policy.}
  \label{fig:s-full-results}
\end{figure*}

Figure~\ref{fig:s-topology} shows the compression effect spatially: fusion removes repeated projected points while retaining category coverage across floors. Figure~\ref{fig:s-cases} then ties three aggregate findings to raw evidence---wider visibility, source-supported fusion, and verifier-controlled recovery---and separately records the physical mapping-query-execution sequence. Figure~\ref{fig:s-ablation} makes the same-response comparison visual across model and FoV controls, whereas Figure~\ref{fig:s-full-results} restores full-validation scale and separates raw single/Top-3, fused single, and fusion-aware recovery. Together, the figures move from map structure, through mechanism audits, to aggregate outcome rather than serving as interchangeable qualitative examples.

\ifdefined\SSTGEmbeddedSupplement
Representative stills in Figures~\ref{fig:s-topology}--\ref{fig:s-full-results} make the qualitative mechanisms directly inspectable in this document. Additional demonstrations are available from the project page linked after the abstract.
\else
Representative stills in Figures~\ref{fig:s-topology}--\ref{fig:s-full-results} make the qualitative mechanisms directly inspectable in this document. The accompanying multimedia contains first-person and top-down navigation videos together with the complete real-robot run. Numeric sources for the tables and aggregate figures are indexed in \path{CLAIM_TO_FILE.md}.
\fi

\section{Physical-System Implementation}

\begin{figure*}[t]
  \centering
  \includegraphics[width=0.98\textwidth]{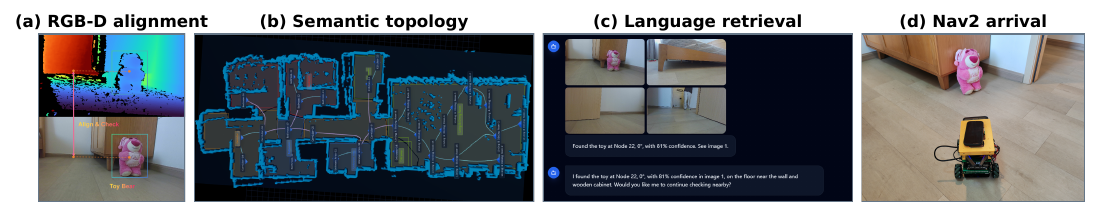}
  \caption{Physical-system realization on the Yahboom X3. (a) Aligned Gemini 336L depth and RGB localize the toy observation. (b) The one-time survey produces a persistent spatial-semantic topology over the mapped environment. (c) The language interface retrieves the toy at Node~22 with 0.81 confidence and presents the supporting views. (d) Nav2 executes the selected target and brings the robot to the toy. These panels document qualitative system integration; SR/SPL results are measured in simulation.}
  \label{fig:s-robot}
\end{figure*}

\ifdefined\SSTGEmbeddedSupplement
The physical platform is a Yahboom X3 mobile robot running ROS~2 Humble \cite{macenski2022ros2}. A Gemini 336L RGB-D camera supplies visual observations, an RPLIDAR S2 supplies planar laser scans, \texttt{slam\_toolbox} \cite{macenski2021slamtoolbox} maintains the occupancy map and localization estimate, and Nav2 \cite{macenski2020marathon2} executes metric goals. Figure~\ref{fig:s-robot} expands the physical sequence shown in the main paper; additional demonstrations are available on the project page.
\else
The physical platform is a Yahboom X3 mobile robot running ROS~2 Humble \cite{macenski2022ros2}. A Gemini 336L RGB-D camera supplies visual observations, an RPLIDAR S2 supplies planar laser scans, \texttt{slam\_toolbox} \cite{macenski2021slamtoolbox} maintains the occupancy map and localization estimate, and Nav2 \cite{macenski2020marathon2} executes metric goals. Figure~\ref{fig:s-robot} expands the physical sequence shown in the main paper; the corresponding run is shown in the separate multimedia submission.
\fi

Perception supplies aligned RGB-D observations, the persistent manager owns the reusable graph shown in Figure~\ref{fig:s-robot}(b), the language interface normalizes the request and exposes its supporting views, and semantic/topological planning dispatches the selected standoff through Nav2. Interaction state retains the remaining candidate list across navigation legs. Quantitative SR/SPL results in the paper come from the simulator benchmark; the physical sequence documents the implemented robot interface.

\ifdefined\SSTGEmbeddedSupplement
\else
\section{Reproducibility Details}

\subsection{Hardware and Software}

Experiments were executed on Ubuntu 24.04.4 LTS with an Intel Core i9-12900KF (24 logical CPUs), 62\,GiB RAM, and an NVIDIA GeForce RTX 3080 Ti with 12\,GiB VRAM. The simulator environment uses Python 3.9, Habitat-Sim 0.3.3, NumPy 1.26.4, PyTorch 2.8.0+cu128, Pillow 11.1.0, imageio 2.37.0, and matplotlib 3.9.1. Hosted API orchestration uses the OpenAI Python client 2.41.0 in a compatible environment. The GPU driver version is 580.159.03.

The physical realization uses a Yahboom X3 base, Gemini 336L RGB-D camera, and RPLIDAR S2. Its ROS~2 Humble software stack uses \texttt{slam\_toolbox} for occupancy mapping and localization, Nav2 for path planning and control, and the SSTG-Nav managers for survey storage, semantic retrieval, candidate sequencing, and arrival verification.

The Qwen full-validation map uses \texttt{Qwen/Qwen2.5-VL-3B-Instruct} in FP16. Hosted models are identified by the model names returned by their endpoints; provider-side weights and serving nondeterminism are not controlled by the authors. Aggregate summaries and per-episode evaluation records are provided in the accompanying artifact.

\subsection{Randomness and Statistics}

Topology seeds are 20260719 plus scene index. Corruption experiments record a seed for every one of 960 conditions. Bootstrap summaries use NumPy's \texttt{default\_rng}; reported primary SPL intervals use seed 0 and 10,000 resamples. Paired post-hoc analysis uses seed 20260719 and 10,000 resamples. VLM decoding is non-streaming; local Qwen uses deterministic generation with sampling disabled. Hosted providers may remain nondeterministic even when request payloads are identical.

\subsection{Artifact Availability}

\path{CLAIM_TO_FILE.md} maps each main-paper table and supplementary analysis to its aggregate summary and episode-level records. Full independent-topology results are provided under \path{evidence/outputs/hm3d_val_uniform/}; target-view controls under \path{evidence/outputs/hm3d_val_oracle_analysis/} and \path{evidence/outputs/hm3d_val_qwen_analysis/}; compact backend, field-of-view, fusion, and density studies under \path{evidence/outputs/hm3d_minival_uniform/}; and corruption analyses under \path{evidence/outputs/analysis/}.

Each headline full-validation condition includes an aggregate JSON summary and an episode-level CSV. Environment specifications, ordered experiment commands, and the implementation modules used for topology construction, RGB-D grounding, fusion, candidate ranking, arrival evaluation, and metric computation are provided under \path{reproduction/}.
\fi

\section{Additional Limitations and Intended Scope}

The main paper evaluates category-level ObjectNav. The natural-language interface is part of the reusable robot system, but relational queries such as ``the vase next to the sofa'' are not represented in HM3D ObjectNav-v2 and are not quantitatively evaluated here. The six-class prompts likewise do not establish open-vocabulary performance beyond the benchmark categories.

The independent map is static. Object motion, furniture rearrangement, appearance change, and map aging are not evaluated. The implemented verifier uses fresh simulator RGB-D and benchmark-category prompts; physical-world calibration, occlusion change, and sensor noise may alter its precision/recall. The oracle-feedback Top-$K$ table remains only a candidate-list ceiling and is never used by the autonomous loop. The simulator uses continuous navmesh shortest paths, so collision recovery, actuator noise, localization drift, camera motion blur, and discrete action limits are absent. GPT-5.4 and MiMo-v2.5 are accessed through hosted endpoints whose weights and future serving behavior are not controlled by the authors. The full independent RGB-D result covers one 1,000-episode HM3D-v2 validation split and one primary semantic backend. Its confidence intervals quantify episode sampling uncertainty, not cross-dataset, model-serving, or real-world variation.

\ifdefined\SSTGEmbeddedSupplement
\let\SSTGMaybeEnd\relax
\else
\clearpage
\raggedbottom
\setlength{\bibsep}{0pt}
\bibliography{aaai2027}
\def\SSTGMaybeEnd{\end{document}}
\fi
\SSTGMaybeEnd

\fi

\bibliography{aaai2027}

@article{anderson2018evaluation,
  title={On Evaluation of Embodied Navigation Agents},
  author={Anderson, Peter and Chang, Angel and Chaplot, Devendra Singh and Dosovitskiy, Alexey and Gupta, Saurabh and Koltun, Vladlen and Kosecka, Jana and Malik, Jitendra and Mottaghi, Roozbeh and Savva, Manolis and Zamir, Amir R.},
  journal={arXiv preprint arXiv:1807.06757},
  year={2018},
  url={https://arxiv.org/abs/1807.06757}
}

@article{batra2020objectnav,
  title={ObjectNav Revisited: On Evaluation of Embodied Agents Navigating to Objects},
  author={Batra, Dhruv and Gokaslan, Aaron and Kembhavi, Aniruddha and Maksymets, Oleksandr and Mottaghi, Roozbeh and Savva, Manolis and Toshev, Alexander and Wijmans, Erik},
  journal={arXiv preprint arXiv:2006.13171},
  year={2020},
  url={https://arxiv.org/abs/2006.13171}
}

@inproceedings{chaplot2020semexp,
  title={Object Goal Navigation Using Goal-Oriented Semantic Exploration},
  author={Chaplot, Devendra Singh and Gandhi, Dhiraj and Gupta, Abhinav and Salakhutdinov, Ruslan},
  booktitle={NeurIPS},
  volume={33},
  pages={4247--4258},
  year={2020},
  url={https://proceedings.neurips.cc/paper/2020/hash/2c75cf2681788adaca63aa95ae028b22-Abstract.html}
}

@inproceedings{huang2023vlmaps,
  title={Visual Language Maps for Robot Navigation},
  author={Huang, Chenguang and Mees, Oier and Zeng, Andy and Burgard, Wolfram},
  booktitle={ICRA},
  pages={10608--10615},
  year={2023},
  doi={10.1109/ICRA48891.2023.10160969},
  url={https://doi.org/10.1109/ICRA48891.2023.10160969}
}

@inproceedings{jatavallabhula2023conceptfusion,
  title={ConceptFusion: Open-Set Multimodal 3D Mapping},
  author={Jatavallabhula, Krishna Murthy and Kuwajerwala, Alihusein and Gu, Qiao and Omama, Mohd and Chen, Tao and Maalouf, Alaa and Li, Shuang and Iyer, Ganesh and Saryazdi, Soroush and Keetha, Nikhil and Tewari, Ayush and Tenenbaum, Joshua B. and de Melo, Celso Miguel and Krishna, Madhava and Paull, Liam and Shkurti, Florian and Torralba, Antonio},
  booktitle={Robotics: Science and Systems},
  year={2023},
  doi={10.15607/RSS.2023.XIX.066},
  url={https://doi.org/10.15607/RSS.2023.XIX.066}
}

@article{kim2022tsgm,
  title={Topological Semantic Graph Memory for Image-Goal Navigation},
  author={Kim, Nuri and Kwon, Obin and Yoo, Hwiyeon and Choi, Yunho and Park, Jeongho and Oh, Songhwai},
  journal={arXiv preprint arXiv:2209.08274},
  year={2022},
  url={https://arxiv.org/abs/2209.08274}
}

@article{peng2026structured,
  title={Structured observation language for efficient and generalizable vision-language navigation},
  author={Peng, Daojie and Ma, Fulong and Ma, Jun},
  journal={arXiv preprint arXiv:2603.27577},
  year={2026}
}

@inproceedings{ko2024care,
  title={Context-Aware Replanning with Pre-Explored Semantic Map for Object Navigation},
  author={Ko, Po-Chen and Su, Hung-Ting and Chen, Ching-Yuan and Yeh, Jia-Fong and Sun, Min and Hsu, Winston H.},
  booktitle={CoRL},
  volume={270},
  pages={4253--4267},
  year={2025},
  url={https://proceedings.mlr.press/v270/ko25b.html}
}

@article{peng2025lovon,
  title={Lovon: Legged open-vocabulary object navigator},
  author={Peng, Daojie and Cao, Jiahang and Zhang, Qiang and Ma, Jun},
  journal={arXiv preprint arXiv:2507.06747},
  year={2025}
}

@article{liu2025toponav,
  title={TopoNav: Topological Graphs as a Key Enabler for Advanced Object Navigation},
  author={Liu, Peiran and Zhang, Qiang and Peng, Daojie and Zhang, Lingfeng and Qin, Yihao and Zhou, Hang and Ma, Jun and Xu, Renjing and Ji, Yiding},
  journal={arXiv preprint arXiv:2509.01364},
  year={2025},
  url={https://arxiv.org/abs/2509.01364}
}

@inproceedings{ramakrishnan2021hm3d,
  title={Habitat-Matterport 3D Dataset ({HM3D}): 1000 Large-Scale 3D Environments for Embodied {AI}},
  author={Ramakrishnan, Santhosh K. and Gokaslan, Aaron and Wijmans, Erik and Maksymets, Oleksandr and Clegg, Alex and Turner, John and Undersander, Eric and Galuba, Wojciech and Westbury, Andrew and Chang, Angel X. and Savva, Manolis and Zhao, Yili and Batra, Dhruv},
  booktitle={NeurIPS Datasets and Benchmarks Track},
  volume={1},
  year={2021},
  url={https://datasets-benchmarks-proceedings.neurips.cc/paper_files/paper/2021/hash/34173cb38f07f89ddbebc2ac9128303f-Abstract-round2.html}
}

@inproceedings{shafiullah2023clipfields,
  title={{CLIP}-Fields: Weakly Supervised Semantic Fields for Robotic Memory},
  author={Shafiullah, Nur Muhammad Mahi and Paxton, Chris and Pinto, Lerrel and Chintala, Soumith and Szlam, Arthur},
  booktitle={Robotics: Science and Systems},
  year={2023},
  doi={10.15607/RSS.2023.XIX.074},
  url={https://doi.org/10.15607/RSS.2023.XIX.074}
}

@inproceedings{takmaz2023openmask3d,
  title={{OpenMask3D}: Open-Vocabulary 3D Instance Segmentation},
  author={Takmaz, Ayca and Fedele, Elisabetta and Sumner, Robert W. and Pollefeys, Marc and Tombari, Federico and Engelmann, Francis},
  booktitle={NeurIPS},
  volume={36},
  pages={68367--68390},
  year={2023},
  url={https://proceedings.neurips.cc/paper_files/paper/2023/hash/d77b5482e38339a8068791d939126be2-Abstract-Conference.html}
}

@article{wang2026consistnav,
  title={ConsistNav: Closing the Action Consistency Gap in Zero-Shot Object Navigation with Semantic Executive Control},
  author={Wang, Haosen and Li, Zhenyang and Zhang, Yinqiang and He, Zongqi and Jiang, Lutao and Li, Kai and Zhao, Yizhou and Fan, Liaoyuan and Hou, Wenjian and Liang, Tingbang and Wen, Yibin and Gu, Defeng},
  journal={arXiv preprint arXiv:2605.09869},
  year={2026},
  url={https://arxiv.org/abs/2605.09869}
}

@inproceedings{wu2024voronav,
  title={VoroNav: Voronoi-Based Zero-Shot Object Navigation with Large Language Model},
  author={Wu, Pengying and Mu, Yao and Wu, Bingxian and Hou, Yi and Ma, Ji and Zhang, Shanghang and Liu, Chang},
  booktitle={ICML},
  volume={235},
  pages={53757--53775},
  year={2024},
  url={https://proceedings.mlr.press/v235/wu24u.html}
}

@inproceedings{yin2024sgnav,
  title={{SG-Nav}: Online 3D Scene Graph Prompting for {LLM}-Based Zero-Shot Object Navigation},
  author={Yin, Hang and Xu, Xiuwei and Wu, Zhenyu and Zhou, Jie and Lu, Jiwen},
  booktitle={NeurIPS},
  volume={37},
  pages={10012--10036},
  year={2024},
  url={https://proceedings.neurips.cc/paper_files/paper/2024/hash/098491b37deebbe6c007e69815729e09-Abstract-Conference.html}
}

@inproceedings{yokoyama2024vlfm,
  title={{VLFM}: Vision-Language Frontier Maps for Zero-Shot Semantic Navigation},
  author={Yokoyama, Naoki and Ha, Sehoon and Batra, Dhruv and Wang, Jiuguang and Bucher, Bernadette},
  booktitle={ICRA},
  pages={42--48},
  year={2024},
  doi={10.1109/ICRA57147.2024.10610712},
  url={https://doi.org/10.1109/ICRA57147.2024.10610712}
}

@inproceedings{zhou2023esc,
  title={{ESC}: Exploration with Soft Commonsense Constraints for Zero-Shot Object Navigation},
  author={Zhou, Kaiwen and Zheng, Kaizhi and Pryor, Connor and Shen, Yilin and Jin, Hongxia and Getoor, Lise and Wang, Xin Eric},
  booktitle={ICML},
  volume={202},
  pages={42829--42842},
  year={2023},
  url={https://proceedings.mlr.press/v202/zhou23r.html}
}

@inproceedings{long2024instructnav,
  title={InstructNav: Zero-Shot System for Generic Instruction Navigation in Unexplored Environment},
  author={Long, Yuxing and Cai, Wenzhe and Wang, Hongcheng and Zhan, Guanqi and Dong, Hao},
  booktitle={CoRL},
  volume={270},
  pages={2049--2060},
  year={2025},
  url={https://proceedings.mlr.press/v270/long25b.html}
}

@article{zhang2025apexnav,
  title={ApexNav: An Adaptive Exploration Strategy for Zero-Shot Object Navigation with Target-Centric Semantic Fusion},
  author={Zhang, Mingjie and Du, Yuheng and Wu, Chengkai and Zhou, Jinni and Qi, Zhenchao and Ma, Jun and Zhou, Boyu},
  journal={IEEE Robotics and Automation Letters},
  volume={10},
  number={11},
  pages={11530--11537},
  year={2025},
  doi={10.1109/LRA.2025.3606388},
  url={https://doi.org/10.1109/LRA.2025.3606388}
}

@article{chabal2025fomnav,
  title={{FOM-Nav}: Frontier-Object Maps for Object Goal Navigation},
  author={Chabal, Thomas and Chen, Shizhe and Ponce, Jean and Schmid, Cordelia},
  journal={arXiv preprint arXiv:2512.01009},
  year={2025},
  url={https://arxiv.org/abs/2512.01009}
}

@article{cai2026intentnav,
  title={IntentNav: Learning Spatial-Visual Object Navigation from Human Demonstrations},
  author={Cai, Yuxin and Li, Zongtai and Wang, Maonan and Bao, Muyi and Zhu, Haokun and Bai, Ruofei and Zhao, Ding and Li, Zirui and Wang, Wenshan and Yau, Wei-Yun and Zhang, Ji and Lv, Chen},
  journal={arXiv preprint arXiv:2606.08029},
  year={2026},
  url={https://arxiv.org/abs/2606.08029}
}

@inproceedings{cao2025cognav,
  title={CogNav: Cognitive Process Modeling for Object Goal Navigation with {LLM}s},
  author={Cao, Yihan and Zhang, Jiazhao and Yu, Zhinan and Liu, Shuzhen and Qin, Zheng and Zou, Qin and Du, Bo and Xu, Kai},
  booktitle={ICCV},
  pages={9550--9560},
  year={2025},
  url={https://openaccess.thecvf.com/content/ICCV2025/html/Cao_CogNav_Cognitive_Process_Modeling_for_Object_Goal_Navigation_with_LLMs_ICCV_2025_paper.html}
}

@inproceedings{zhang2025fbn,
  title={Function-centric Bayesian Network for Zero-Shot Object Goal Navigation},
  author={Zhang, Sixian and Yu, Xinyao and Song, Xinhang and Wang, Yiyao and Jiang, Shuqiang},
  booktitle={ICCV},
  pages={19535--19545},
  year={2025},
  url={https://openaccess.thecvf.com/content/ICCV2025/html/Zhang_Function-centric_Bayesian_Network_for_Zero-Shot_Object_Goal_Navigation_ICCV_2025_paper.html}
}

@inproceedings{wang2026trajrag,
  title={TrajRAG: Retrieving Geometric-Semantic Experience for Zero-Shot Object Navigation},
  author={Wang, Yiyao and Zhang, Sixian and Zhang, Keming and Song, Xinhang and Du, Songjie and Jiang, Shuqiang},
  booktitle={CVPR},
  pages={15166--15176},
  year={2026},
  url={https://openaccess.thecvf.com/content/CVPR2026/html/Wang_TrajRAG_Retrieving_Geometric-Semantic_Experience_for_Zero-Shot_Object_Navigation_CVPR_2026_paper.html}
}

@inproceedings{yu2023l3mvn,
  title={{L3MVN}: Leveraging Large Language Models for Visual Target Navigation},
  author={Yu, Bangguo and Kasaei, Hamidreza and Cao, Ming},
  booktitle={IROS},
  pages={3554--3560},
  year={2023},
  doi={10.1109/IROS55552.2023.10342512},
  url={https://doi.org/10.1109/IROS55552.2023.10342512}
}

@inproceedings{yuan2024gamap,
  title={{GAMap}: Zero-Shot Object Goal Navigation with Multi-Scale Geometric-Affordance Guidance},
  author={Yuan, Shuaihang and Huang, Hao and Hao, Yu and Wen, Congcong and Tzes, Anthony and Fang, Yi},
  booktitle={NeurIPS},
  volume={37},
  pages={39386--39408},
  year={2024},
  url={https://proceedings.neurips.cc/paper_files/paper/2024/hash/459d93dad139eb084c365d40a57eada3-Abstract-Conference.html}
}

@inproceedings{campari2022reusable,
  title={Online Learning of Reusable Abstract Models for Object Goal Navigation},
  author={Campari, Tommaso and Lamanna, Leonardo and Traverso, Paolo and Serafini, Luciano and Ballan, Lamberto},
  booktitle={CVPR},
  pages={14870--14879},
  year={2022},
  url={https://openaccess.thecvf.com/content/CVPR2022/html/Campari_Online_Learning_of_Reusable_Abstract_Models_for_Object_Goal_Navigation_CVPR_2022_paper.html}
}

@inproceedings{ramakrishnan2022poni,
  title={{PONI}: Potential Functions for ObjectGoal Navigation with Interaction-Free Learning},
  author={Ramakrishnan, Santhosh Kumar and Chaplot, Devendra Singh and Al-Halah, Ziad and Malik, Jitendra and Grauman, Kristen},
  booktitle={CVPR},
  pages={18890--18900},
  year={2022},
  url={https://openaccess.thecvf.com/content/CVPR2022/html/Ramakrishnan_PONI_Potential_Functions_for_ObjectGoal_Navigation_With_Interaction-Free_Learning_CVPR_2022_paper.html}
}

@inproceedings{majumdar2022zson,
  title={{ZSON}: Zero-Shot Object-Goal Navigation using Multimodal Goal Embeddings},
  author={Majumdar, Arjun and Aggarwal, Gunjan and Devnani, Bhavika and Hoffman, Judy and Batra, Dhruv},
  booktitle={NeurIPS},
  volume={35},
  pages={32340--32352},
  year={2022},
  url={https://proceedings.neurips.cc/paper_files/paper/2022/hash/d0b8f0c8f79d3a621af945cafb669f4b-Abstract-Conference.html}
}

@inproceedings{gadre2023cow,
  title={{CoWs} on Pasture: Baselines and Benchmarks for Language-Driven Zero-Shot Object Navigation},
  author={Gadre, Samir Yitzhak and Wortsman, Mitchell and Ilharco, Gabriel and Schmidt, Ludwig and Song, Shuran},
  booktitle={CVPR},
  pages={23171--23181},
  year={2023},
  url={https://openaccess.thecvf.com/content/CVPR2023/html/Gadre_CoWs_on_Pasture_Baselines_and_Benchmarks_for_Language-Driven_Zero-Shot_Object_CVPR_2023_paper.html}
}

@inproceedings{cai2024pixnav,
  title={Bridging Zero-Shot Object Navigation and Foundation Models through Pixel-Guided Navigation Skill},
  author={Cai, Wenzhe and Huang, Siyuan and Cheng, Guangran and Long, Yuxing and Gao, Peng and Sun, Changyin and Dong, Hao},
  booktitle={ICRA},
  pages={5228--5234},
  year={2024},
  url={https://ieeexplore.ieee.org/document/10610499/}
}

@inproceedings{zhang2024imagine,
  title={Imagine Before Go: Self-Supervised Generative Map for Object Goal Navigation},
  author={Zhang, Sixian and Yu, Xinyao and Song, Xinhang and Wang, Xiaohan and Jiang, Shuqiang},
  booktitle={CVPR},
  pages={16414--16425},
  year={2024},
  url={https://openaccess.thecvf.com/content/CVPR2024/html/Zhang_Imagine_Before_Go_Self-Supervised_Generative_Map_for_Object_Goal_Navigation_CVPR_2024_paper.html}
}

@inproceedings{yin2025unigoal,
  title={UniGoal: Towards Universal Zero-Shot Goal-Oriented Navigation},
  author={Yin, Hang and Xu, Xiuwei and Zhao, Linqing and Wang, Ziwei and Zhou, Jie and Lu, Jiwen},
  booktitle={CVPR},
  pages={19057--19066},
  year={2025},
  url={https://openaccess.thecvf.com/content/CVPR2025/html/Yin_UniGoal_Towards_Universal_Zero-shot_Goal-oriented_Navigation_CVPR_2025_paper.html}
}

@article{nie2025wmnav,
  title={{WMNav}: Integrating Vision-Language Models into World Models for Object Goal Navigation},
  author={Nie, Dujun and Guo, Xianda and Duan, Yiqun and Zhang, Ruijun and Chen, Long},
  journal={arXiv preprint arXiv:2503.02247},
  year={2025},
  url={https://arxiv.org/abs/2503.02247}
}

@inproceedings{savva2019habitat,
  title={{Habitat}: A Platform for Embodied {AI} Research},
  author={Savva, Manolis and Kadian, Abhishek and Maksymets, Oleksandr and Zhao, Yili and Wijmans, Erik and Jain, Bhavana and Straub, Julian and Liu, Jia and Koltun, Vladlen and Malik, Jitendra and Parikh, Devi and Batra, Dhruv},
  booktitle={ICCV},
  pages={9339--9347},
  year={2019},
  doi={10.1109/ICCV.2019.00943},
  url={https://doi.org/10.1109/ICCV.2019.00943}
}

@inproceedings{yadav2023hm3dsem,
  title={{Habitat-Matterport 3D Semantics Dataset}},
  author={Yadav, Karmesh and Ramrakhya, Ram and Ramakrishnan, Santhosh Kumar and Gervet, Theo and Turner, John and Gokaslan, Aaron and Maestre, Noah and Chang, Angel Xuan and Batra, Dhruv and Savva, Manolis and Clegg, Alexander William and Chaplot, Devendra Singh},
  booktitle={CVPR},
  pages={4927--4936},
  year={2023},
  doi={10.1109/CVPR52729.2023.00477},
  url={https://doi.org/10.1109/CVPR52729.2023.00477}
}

@misc{habitatchallenge2023,
  title={{Habitat Challenge 2023}},
  author={Yadav, Karmesh and Krantz, Jacob and Ramrakhya, Ram and Ramakrishnan, Santhosh Kumar and Yang, Jimmy and Wang, Austin and Turner, John and Gokaslan, Aaron and Berges, Vincent-Pierre and Mottaghi, Roozbeh and Maksymets, Oleksandr and Chang, Angel X. and Savva, Manolis and Clegg, Alexander and Chaplot, Devendra Singh and Batra, Dhruv},
  howpublished={\url{https://aihabitat.org/challenge/2023/}},
  year={2023}
}

@article{bai2025qwen25vl,
  title={{Qwen2.5-VL} Technical Report},
  author={Bai, Shuai and Chen, Keqin and Liu, Xuejing and Wang, Jialin and Ge, Wenbin and Song, Sibo and Dang, Kai and Wang, Peng and Wang, Shijie and Tang, Jun and Zhong, Humen and Zhu, Yuanzhi and Yang, Mingkun and Li, Zhaohai and Wan, Jianqiang and Wang, Pengfei and Ding, Wei and Fu, Zheren and Xu, Yiheng and Ye, Jiabo and Zhang, Xi and Xie, Tianbao and Cheng, Zesen and Zhang, Hang and Yang, Zhibo and Xu, Haiyang and Lin, Junyang},
  journal={arXiv preprint arXiv:2502.13923},
  year={2025},
  url={https://arxiv.org/abs/2502.13923}
}

@misc{openai2026gpt54,
  title={Introducing {GPT-5.4}},
  author={{OpenAI}},
  year={2026},
  howpublished={\url{https://openai.com/index/introducing-gpt-5-4/}}
}

@misc{mimov25,
  title={{MiMo-V2.5}},
  author={{Xiaomi MiMo Team}},
  year={2026},
  howpublished={\url{https://huggingface.co/collections/XiaomiMiMo/mimo-v25}}
}

@article{macenski2022ros2,
  title={Robot Operating System 2: Design, Architecture, and Uses in the Wild},
  author={Macenski, Steven and Foote, Tully and Gerkey, Brian and Lalancette, Chris and Woodall, William},
  journal={Science Robotics},
  volume={7},
  number={66},
  pages={eabm6074},
  year={2022},
  doi={10.1126/scirobotics.abm6074},
  url={https://doi.org/10.1126/scirobotics.abm6074}
}

@inproceedings{macenski2020marathon2,
  title={The Marathon 2: A Navigation System},
  author={Macenski, Steve and Mart{\'i}n, Francisco and White, Ruffin and Gin{\'e}s Clavero, Jonatan},
  booktitle={IROS},
  pages={2718--2725},
  year={2020},
  doi={10.1109/IROS45743.2020.9341207},
  url={https://doi.org/10.1109/IROS45743.2020.9341207}
}

@article{macenski2021slamtoolbox,
  title={{SLAM Toolbox}: {SLAM} for the Dynamic World},
  author={Macenski, Steve and Jambrecic, Ivona},
  journal={Journal of Open Source Software},
  volume={6},
  number={61},
  pages={2783},
  year={2021},
  doi={10.21105/joss.02783},
  url={https://doi.org/10.21105/joss.02783}
}

\end{document}